\documentclass{article}

    \PassOptionsToPackage{numbers, compress}{natbib}
 \usepackage[preprint]{neurips_2026}

\usepackage[utf8]{inputenc} 
\usepackage[T1]{fontenc}    
\usepackage{hyperref}       
\usepackage{url}            
\usepackage{booktabs}       
\usepackage{amsfonts}       
\usepackage{nicefrac}       
\usepackage{microtype}      
\usepackage{xcolor}         
\usepackage{amsmath}
\usepackage{graphicx}

\title{ReportQA: QA-Based Radiology Report Evaluation}

\author{%
  Yiming Shi\textsuperscript{1\footnotemark[1]} \quad Shaoshuai Yang\textsuperscript{1\footnotemark[1]} \quad Xi Chen\textsuperscript{1} \quad Haolin Li\textsuperscript{1} \quad Hengyu Zhang\textsuperscript{1} \\
  \textbf{Che Jiang\textsuperscript{1}} \quad \textbf{Kaiwen Wang\textsuperscript{1}} \quad \textbf{Xun Zhu\textsuperscript{1}} \quad \textbf{Dong Xie\textsuperscript{4}} \quad \textbf{Fei Wang\textsuperscript{4}} \\
  \textbf{Dejing Dou\textsuperscript{4}} \quad \textbf{Miao Li\textsuperscript{1\footnotemark[2]}} \quad \textbf{Ji Wu\textsuperscript{1, 2, 3\footnotemark[2]}} \\
  \textsuperscript{1} Department of Electronic Engineering, Tsinghua University \\
  \textsuperscript{2} College of AI, Tsinghua University \\
  \textsuperscript{3} Beijing National Research Center for Information Science and Technology \\
  \textsuperscript{4} Beijing Electronic Digital \& Intelligence \\
  \texttt{\{sym23,yangss24\}@mails.tsinghua.edu.cn} \\
  \texttt{\{miao-li, wuji\_ee\}@tsinghua.edu.cn} \\
}

\begin{document}

\maketitle

\renewcommand{\thefootnote}{\fnsymbol{footnote}}
\footnotetext[1]{Equal contribution}
\footnotetext[2]{Corresponding authors: Ji Wu and Miao Li}
\footnotetext[3]{HuggingFace: \url{https://huggingface.co/datasets/shiym2000/ReportQA}}
\footnotetext[4]{GitHub: \url{https://github.com/MSIIP/ReportQA}}

\begin{abstract}
  Radiology report evaluation is essential for advancing automated radiology report generation. Common natural language generation metrics have limited clinical relevance. Clinical efficacy (CE) metrics evaluate important medical findings, but focus mainly on presence and cover only a limited set of entities. Due to heavy reliance on manual annotations, it is difficult for CE metrics to extend clinical entities or attributes. In clinical practice, radiology reports serve as a medium for information transfer. Clinicians use them to perform downstream diagnostic tasks without directly inspecting images. Based on this insight, we propose \textbf{ReportQA}, a clinical-related and flexible radiology report evaluation framework, supporting \textbf{detailed quantitative analysis} of radiology report generation systems. We first collect datasets covering multiple imaging modalities and anatomical regions. We then construct knowledge trees of clinical entities and attributes with radiologist guidance, and use large language models (LLMs) to extract structured information from free-form reports. Next, we generate QA pairs from predefined templates and apply quality control through self-filtering and report-based filtering. During evaluation, the report is treated as context, and an LLM acts as a judge model to answer the QA pairs. Based on the resulting QA accuracy, we introduce \textbf{QAScore} metric. Compared with existing metrics, QAScore shows better alignment with radiologist judgments. Experiments on multiple state-of-the-art vision-language models reveal that current report-based inference paradigms struggle to learn fine-grained clinical representations and exhibit strong negative prior biases. In contrast, \textbf{question-driven inference} provides a more effective alternative. For reproducibility and extensibility, we release the knowledge trees, structured reports, and QA pairs\footnotemark[3], along with the pipeline code\footnotemark[4] for QA construction and evaluation.
\end{abstract}

\section{Introduction}

Medical imaging is indispensable in routine clinical practice~\citep{mia}. This has motivated growing interest in automated radiology report generation (RRG), which can improve clinical efficiency and support consistent report quality. However, progress in RRG is constrained by the lack of reliable radiology report evaluation (RRE) frameworks. Existing RRE frameworks fail to provide meaningful feedback on clinical correctness, which limits the effective development of RRG systems.

Existing RRE frameworks typically combine natural language generation (NLG) metrics with clinical efficacy (CE) metrics. However, they remain limited in clinical relevance and flexibility. NLG metrics, such as BLEU~\citep{bleu}, mainly measure lexical overlap between generated and ground-truth reports. This evaluation paradigm is misaligned with clinical requirements. In practice, a report is clinically useful not because it is lexically similar to a reference report, but because it accurately describes key clinical entities and their associated attributes.

CE metrics, such as CheXbert~\citep{chexbert}, partially address this issue by evaluating the presence of selected clinical entities. However, they remain limited to coarse-grained existence modeling. They fail to capture fine-grained attributes such as location, shape and size. Moreover, CE metrics rely on supervised models trained with substantial manual annotations. This makes them difficult to extend to new entities or attributes and thus limiting detailed quantitative analysis of RRG system.

In clinical practice, radiology reports serve as a medium for information transfer~\citep{rr1,rr2}. Clinicians use report content instead of raw imaging for a wide range of downstream diagnostic tasks. Motivated by this, we propose to evaluate reports by treating them as context for question answering (QA). This design simulates how clinicians interact with reports in real-world settings. The resulting QA-based evaluation framework better aligns RRE with the clinical use of RRG systems and provides a more faithful measure of report utility.

First, we collect diverse radiology report datasets covering multiple imaging modalities and anatomical regions. To better align with real-world clinical scenarios, we transform raw free-form reports into a structured format using dataset-specific knowledge trees designed by radiologists. The structured reports are then converted into QA pairs. We use large language models (LLMs) for quality control. During evaluation, each report is treated as contextual input, and an LLM is used as the judge model to answer the corresponding QA pairs. Based on the LLM responses, we introduce QAScore metric.

Compared with existing RRE metrics, QAScore shows stronger alignment with radiologist judgments. Furthermore, through a comprehensive evaluation of state-of-the-art (SOTA) vision-language models (VLMs), we find that current report-based inference paradigms struggle to learn fine-grained clinical representations and exhibit strong negative prior biases. In contrast, question-driven inference provides a more promising alternative. Our main contributions are summarized as follows:

\begin{enumerate}
\item We propose \textbf{ReportQA}, a QA-based radiology report evaluation framework, supporting detailed quantitative analysis of RRG systems. Additionally, we introduce \textbf{QAScore}, which is an aggregated metric based on the judge model responses.
\item ReportQA reveals \textbf{hidden failure modes} that are invisible to common report-level metrics. Experiments indicate that existing report-based inference paradigms struggle to learn fine-grained clinical information and exhibit strong negative prior biases. In contrast, \textbf{question-driven inference} provides a more effective alternative.
\item We release all knowledge trees, structured reports and QA pairs, along with the pipeline code for QA construction and evaluation. Users can plug in stronger LLMs to improve QA quality and apply the pipeline to more datasets.
\end{enumerate}

\section{Related Work}

\subsection{Radiology report evaluation}

RRE frameworks can be broadly categorized into NLG metrics and CE metrics. Among NLG metrics, BLEU~\citep{bleu} measures n-gram precision between generated and reference texts, ROUGE~\citep{rouge} focuses on recall-oriented overlap, and METEOR~\citep{meteor} incorporates synonym matching and alignment-based scoring. Despite their widespread use, NLG metrics primarily rely on surface-level lexical similarity and fail to capture the accuracy of clinically relevant information.

CE-based metrics attempt to address this limitation by explicitly evaluating clinical entities. CheXbert~\citep{chexbert} is a BERT-based classifier trained to predict the presence of predefined clinical findings from reports, while RadGraph~\citep{radgraph} extracts structured entities and relations to enable more fine-grained evaluation. However, these approaches depend heavily on manually annotated data and predefined label spaces, making them difficult to extend to new clinical entities or attributes.

Recently, several advanced RRE metrics have been proposed. For example, GREEN~\citep{green} leverages LLMs to identify and explain clinical errors in reports, providing scores that better align with radiologist preferences. However, its evaluation operates primarily at a holistic semantic level and lacks systematic modeling of fine-grained clinical entities. Similarly, RaTEScore~\citep{ratescore} introduces entity-level semantic modeling by combining named entity recognition with embedding-based similarity measures to improve consistency assessment of clinically relevant information. Nevertheless, it relies on predefined entity categories and fixed semantic spaces, which constrain its expressiveness and limit flexibility to more complex or diverse clinical scenarios.

In contrast, our QA-based RRE framework is grounded in the fundamental role of radiology reports as clinical information media. It combines clinical relevance and flexibility, enabling detailed quantitative analysis of RRG systems.

\begin{figure}
  \centering
  \includegraphics[width=1\linewidth]{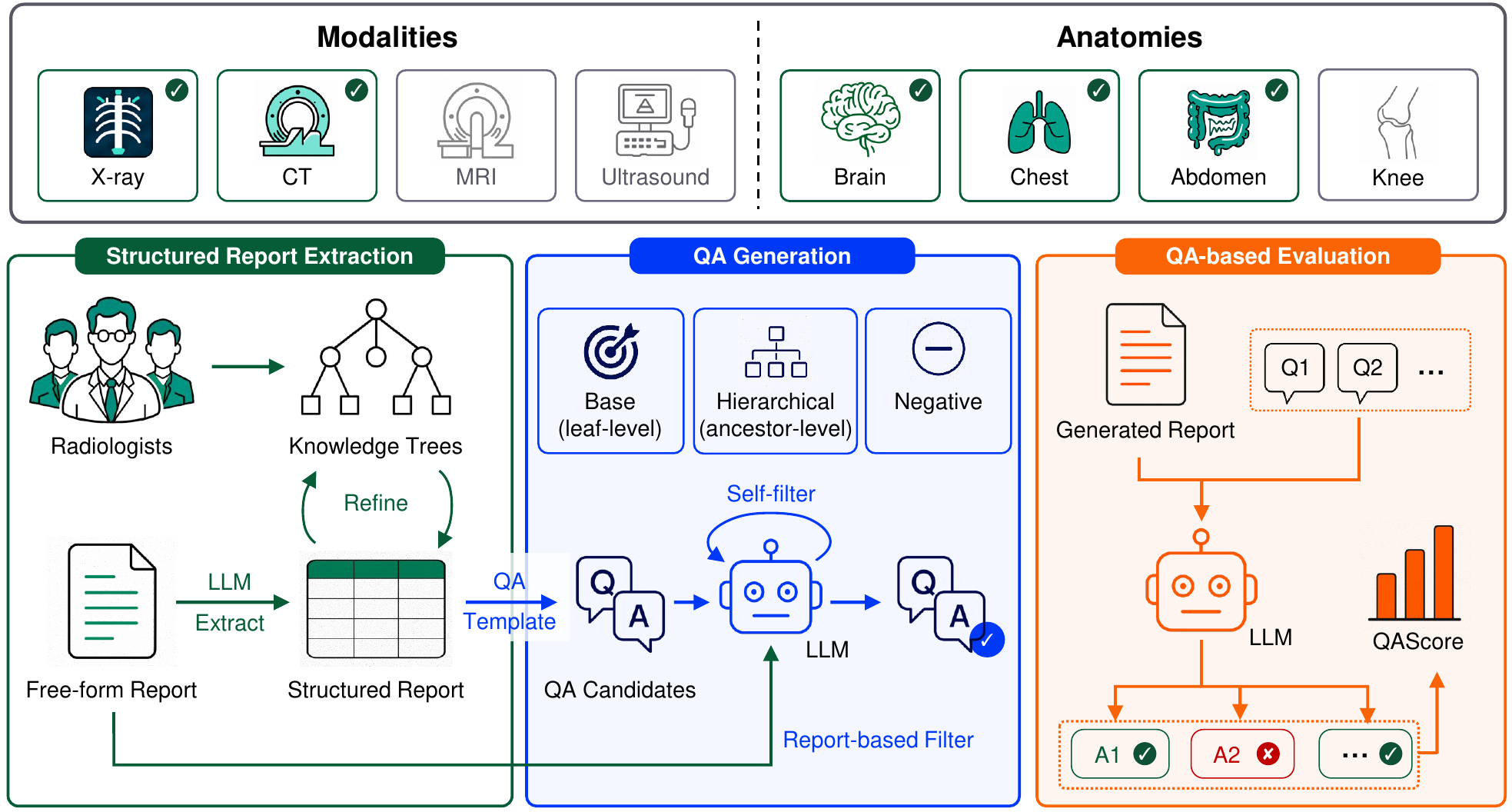}
  \caption{Overview framework of ReportQA. First, we collect datasets covering multiple imaging modalities and anatomical regions. Second, free-form reports are transformed into a structured format based on radiologist-defined knowledge trees. Third, QA pairs are generated via predefined templates and refined through self-filtering and report-based filtering, resulting in three categories of QA: \textit{base (leaf-level)}, \textit{hierarchical (ancestor-level)}, and \textit{negative}. Finally, the generated report is treated as context, and an LLM answers filtered questions. LLM responses are used to calculate QAScore.}
  \label{fig:overall}
\end{figure}

\subsection{Medical visual question answering}

In the medical domain, visual question answering (VQA) provides a unified interface for diverse tasks, including disease diagnosis~\citep{dd}, lesion localization~\citep{ll}, and RRG. PMC-VQA~\citep{pmcvqa} is constructed from the Open Access subset of PubMed Central, covering multiple imaging modalities and anatomical regions. Similarly, OmniMedVQA~\citep{omnimedvqa} and GMAI-MMBench~\citep{gmaimmbench} aggregate data from multiple public datasets to improve coverage across modalities, clinical specialties, and tasks.

However, these datasets are not directly derived from raw clinical workflows, which introduces a potential gap between benchmark performance and real-world applicability. In addition, most existing VQA datasets are limited to 2D images. 3D VQA resources remain scarce: M3D-VQA~\citep{m3d} is constructed from Radiopaedia but raises potential licensing concerns, while CT-RATE~\citep{ctrate} provides a VQA subset that lacks a systematic and quality-controlled construction pipeline, resulting in relatively simple queries that may not meet the demands of current VLMs.

To address these limitations, we construct a high-quality medical VQA dataset by leveraging widely used radiology report datasets across both 2D and 3D imaging. Our approach employs a well-designed pipeline to extract structured reports and generate clinically grounded QA pairs, bridging the gap between real-world clinical data and VQA evaluation.

\section{Method}

\begin{figure}
  \centering
  \includegraphics[width=1\linewidth]{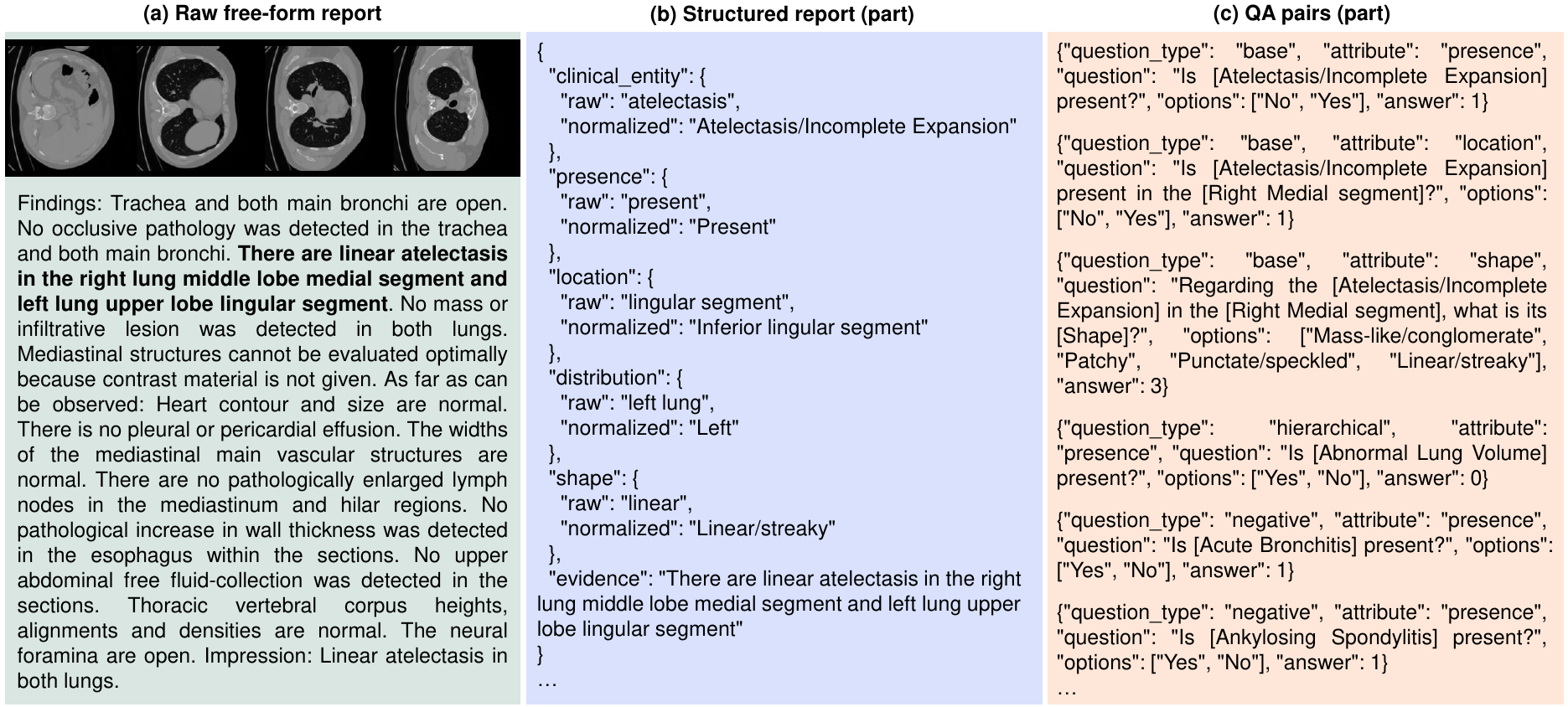}
  \caption{An example from CT-RATE~\citep{ctrate}, including raw free-form report, structured report, and QA pairs. The structured report and QA pairs are simplified and shown only partially for visualization.}
  \label{fig:example}
\end{figure}

As shown in Figure~\ref{fig:overall}, we construct ReportQA through a four-stage pipeline. First, we collect datasets consisting of free-form radiology reports. Second, we transform these reports into a structured format using dataset-specific knowledge trees, which encode clinically relevant entities and their attributes. Third, we generate QA pairs from the structured reports. Finally, we evaluate reports by answering filtered QA pairs, with comprehensive performance quantified via proposed QAScore.

\subsection{Dataset preparation}

ReportQA is constructed from diverse radiology report datasets spanning multiple imaging modalities and anatomical regions, including chest X-ray data from MIMIC-CXR~\citep{mimiccxr}, brain CT data from CTRG-Brain~\citep{ctrg}, chest CT data from CT-RATE~\citep{ctrate}, and abdominal CT data from AMOS-MM~\citep{amos}.

For 2D imaging datasets, images are provided in JPEG format, such as MIMIC-CXR-JPG~\citep{mimiccxrjpg}, while 3D volumes are standardized to the NIfTI format to ensure consistency across datasets. ReportQA is constructed based on the official validation or test splits whenever available. For datasets without predefined splits, such as CTRG-Brain, we adopt a reproducible 4:1 train-validation split.

\subsection{Structured report extraction}

First, radiologists construct two dataset-specific knowledge trees based on dataset metadata, including imaging modality and anatomical region. One knowledge tree models clinical entities, consisting of findings and diagnoses, where findings are primarily extracted from the Findings section of reports and diagnoses from the Impression section. The other tree defines a comprehensive set of clinical attributes, covering 15 dimensions: \textit{presence}, \textit{location}, \textit{distribution}, \textit{number}, \textit{dimension}, \textit{density}, \textit{shape}, \textit{margin}, \textit{enhancement}, \textit{internal features}, \textit{secondary effects}, \textit{severity}, \textit{chronicity}, \textit{clinical score}, and \textit{certainty}. Detailed examples of knowledge trees are provided in Appendix \ref{sec:prompt}.

We then employ LLMs to extract structured information from raw free-form reports. To improve consistency and coverage, we adopt a two-stage extraction process. In the first stage, LLMs generate preliminary structured outputs. These outputs are subsequently used to refine the knowledge trees by identifying missing or ambiguous nodes. In the second stage, the refined knowledge trees are used to remap the extracted entities and attributes, ensuring that all structured information is aligned with predefined nodes. Invalid entities and attributes that cannot be mapped to the knowledge trees are filtered out, resulting in a controlled and standardized representation. Figure~\ref{fig:example} shows the example of structured reports. The prompts used for structured extraction are provided in Appendix \ref{sec:prompt}.

\subsection{QA generation}
\label{sec:qageneration}

\begin{table}
  \caption{Statistics of QA construction across datasets. After filtering, a high QA density is maintained, demonstrating that the filtering strategy effectively improves quality without sacrificing coverage.}
  \label{tab:stat}
  \centering
  \begin{tabular}{lccc}
    \toprule
    Dataset & \#Reports & \#QAs (filtered) & QA density \\
    \midrule
    CTRG-Brain~\citep{ctrg} & 2001 & 182136 & 91.0 \\
    CT-RATE~\citep{ctrate} & 3039 & 298654 & 98.3 \\
    AMOS-MM~\citep{amos} & 400 & 25433 & 63.6 \\
    MIMIC-CXR~\citep{mimiccxr} & 1417 & 153821 & 108.6 \\
    \bottomrule
  \end{tabular}
\end{table}

Given the structured reports, we construct \textit{base (leaf-level)} multiple-choice QA pairs using predefined templates that systematically cover different attributes of clinical entities. As shown in Figure~\ref{fig:example}, each QA instance targets a specific (entity, attribute) combination.

To account for \textit{hierarchical (ancestor-level)} clinical entities, we incorporate ontology-aware question generation. In practice, entities may not be described at the most fine-grained level (leaf-level). For example, if the ground-truth report contains diffuse brain atrophy while the generated report predicts brain atrophy, the prediction can be considered partially correct. To reflect this, we additionally generate QA pairs for ancestor nodes of entities based on hierarchy in the clinical entity tree.

We further construct \textit{negative} QA pairs for clinical entities that are absent in the ground-truth report, focusing on the presence attribute. To mitigate overconfident guessing by LLM-based evaluators when evidence is missing, we include an "Insufficient information" option in the answer choices.

Finally, we employ an LLM as the judge model to perform quality control over QA pairs through a two-stage filtering process. First, we directly evaluate QA pairs without providing report context and remove those that can be correctly answered without contextual information. Second, we re-evaluate QA pairs using the ground-truth report as context and filter out instances with low accuracy. This process ensures that remaining QA pairs are both context-dependent and reliable for evaluation.

After self-filtering and report-based filtering, approximately 660K QAs remain for 6857 radiology reports, averaging nearly 100 high-quality QAs per report. Detailed statistics are provided in Table~\ref{tab:stat}.

\subsection{QA-based evaluation}

Following a procedure similar to QA filtering in Section \ref{sec:qageneration}, we employ an LLM as the judge model to evaluate reports generated by RRG systems. Specifically, the judge model answers filtered QA pairs based on generated reports. The overall performance is quantified using QA accuracy. The framework enables fine-grained analysis by decomposing performance across different question types, allowing for a more detailed understanding of model behavior.

Additionally, we propose an overall metric, termed \textbf{QAScore}, to jointly evaluate a generated report's ability to preserve positive clinical entities and to avoid introducing spurious entities. Specifically, \textit{base (leaf-level)} and \textit{hierarchical (ancestor-level)} questions are used to measure positive consistency, defined as:

\begin{equation}
\text{Score}_{\text{pos}} = \frac{1}{N_{\text{pos}}} \sum_{i=1}^{N_{\text{pos}}} \text{Acc}_i,
\end{equation}

where $N_{\text{pos}}$ denotes the total number of \textit{base (leaf-level)} and \textit{hierarchical (ancestor-level)} questions, and $\text{Acc}_i \in \{0,1\}$ indicates whether the $i$-th question is answered correctly. Notably, if the presence question of a positive entity is answered incorrectly, all associated questions for that entity are considered incorrect, as entity existence is a prerequisite for subsequent attribute reasoning.

\textit{Negative} questions are used to assess whether the generated report introduces entities absent from the ground-truth report. Due to the typically large number of negative questions, accuracy alone may be dominated by trivial correct answers and fail to reflect clinically critical false positives. Therefore, we focus on the false positive rate for negative questions, denoted as $\text{FPR}_{\text{neg}}$, which measures the proportion of negative entities incorrectly predicted as present. The corresponding negative score is defined as:

\begin{equation}
\text{Score}_{\text{neg}} = \exp(-\lambda \cdot \text{FPR}_{\text{neg}}),
\end{equation}

where $\lambda$ controls the sensitivity to false positive errors. The overall report-level score is defined as the harmonic mean of $\text{Score}_{\text{pos}}$ and $\text{Score}_{\text{neg}}$:

\begin{equation}
\text{QAScore} = \frac{2 \cdot \text{Score}_{\text{pos}} \cdot \text{Score}_{\text{neg}}}{\text{Score}_{\text{pos}} + \text{Score}_{\text{neg}}}.
\end{equation}

Finally, we report the average QAScore across all reports, assigning equal weight to each case to avoid bias toward reports with a larger number of questions.

\section{Experiment}

\subsection{Implementation details}

For structured report extraction, we employ DeepSeek-V3.2~\citep{deepseekv32} API. For QA filtering and evaluation, we deploy locally a judge model, Qwen3.5-27B, with vLLM~\citep{vllm} on a single A800 (80GB) GPU.

We evaluate a diverse set of SOTA VLMs, including both proprietary API-based VLMs and open-source general~\citep{internvl35} and medical~\citep{hulumed,lingshu,medgemma15,radfm,ctrate} VLMs. We adopt the official inference protocols released with each model if available. Otherwise, we perform inference using SWIFT~\citep{msswift} or TRL~\citep{trl}. All open-source model inference experiments are conducted on a single A800 (80GB) GPU, except for CT-CHAT (70B), which required two A800 GPUs.

For medical image preprocessing, we follow the recommended configurations provided by each model. For API-based VLMs without explicit guidelines, we uniformly sample three slices at equal intervals for inference. For VLMs lacking predefined settings, 2D images are resized to $512\times512$, and 3D volumes are uniformly resized to 16 slices along the depth dimension.

\subsection{QAScore better aligns with radiologist judgments}

\begin{table}
  \caption{Correlation between automatic metrics and \textit{total\_errors}. QAScore shows the strongest correlation with human-annotated errors, indicating the best alignment with radiologist judgments.}
  \label{tab:metric}
  \centering
  \begin{tabular}{lccc}
    \toprule
    Metric & $|\text{Pearson}|$ & $|\text{Spearman}|$ & $|\text{Kendall}|$ \\
    \midrule
    BLEU-4~\citep{bleu} & 0.1190 & 0.0964 & 0.0728 \\
    RadGraph F1~\citep{radgraph} & 0.1743 & 0.1692 & 0.1238 \\
    BLEU-2~\citep{bleu} & 0.1851 & 0.1742 & 0.1265 \\
    RadCliQ~\citep{radcliq} & 0.2017 & 0.2153 & 0.1586 \\
    BERTScore~\citep{bertscore} & 0.2575 & 0.2525 & 0.1901 \\
    RaTEScore~\citep{ratescore} & 0.3539 & 0.3551 & 0.2651 \\
    CheXbert~\citep{chexbert} & 0.3870 & 0.3779 & 0.2896 \\
    GREEN~\citep{green} & 0.4194 & 0.4022 & 0.3148 \\
    QAScore & \textbf{0.4507} & \textbf{0.4612} & \textbf{0.3910} \\
    \bottomrule
  \end{tabular}
\end{table}

To evaluate the alignment between QAScore and radiologist judgments, we conduct a correlation study on the RadEvalX~\citep{radevalx} dataset. RadEvalX is designed for chest X-ray report evaluation and contains 100 report pairs, each consisting of a ground-truth report from the IU-Xray~\citep{iuxray} dataset, a generated report from the M2Tr~\citep{m2tr} model, and annotations from two board-certified radiologists covering two types of errors across eight error categories. We aggregate the two error types into a single scalar, \textit{total\_errors}, for each report, which serves as the human reference signal.

For automatic evaluation, we include standard metrics provided by RadEvalX, including BLEU~\citep{bleu}, BERTScore~\citep{bertscore}, CheXbert~\citep{chexbert}, RadGraph F1~\citep{radgraph}, and RadCliQ~\citep{radcliq}, and further extend the comparison with two recent radiology-specific metrics, GREEN~\citep{green} and RaTEScore~\citep{ratescore}.

To comprehensively assess alignment with radiologist judgments, we compute Pearson, Spearman, and Kendall correlation coefficients between each metric and \textit{total\_errors}. As shown in Table~\ref{tab:metric}, QAScore achieves the highest absolute correlation across all three measures, with $\text{Pearson}=0.4507$, $\text{Spearman}=0.4612$, and $\text{Kendall}=0.3910$, outperforming GREEN, RaTEScore, and other baseline metrics. These results demonstrate that QAScore more faithfully reflects radiologist judgments of report errors and provides a more reliable measure of clinical quality.

\subsection{Proprietary VLMs better, while 3D datasets remain challenging}

\begin{table}
  \caption{ReportQA results of proprietary VLMs, open-source general VLMs, and open-source medical VLMs across different datasets. \textbf{Bold} indicates the best performance among all models, while \underline{underlined} values denote the best performance within each model category.}
  \label{tab:overall}
  \centering
  \begin{tabular}{lcccc}
    \toprule
    Model & CTRG-Brain & CT-RATE & AMOS-MM & MIMIC-CXR \\
    \midrule
    \multicolumn{5}{l}{\textit{Proprietary VLMs}} \\
    \cmidrule{1-5}
    GPT-5.4 & \underline{\textbf{0.4536}} & \underline{\textbf{0.4702}} & 0.2165 & \underline{\textbf{0.6126}} \\
    Gemini 3.1 Pro & 0.4271 & 0.2009 & 0.0837 & 0.5055 \\
    Claude Opus 4.6 & 0.1912 & 0.2597 & \underline{\textbf{0.3125}} & 0.4995 \\
    \cmidrule{1-5}
    \multicolumn{5}{l}{\textit{Open-source VLMs (general)}} \\
    \cmidrule{1-5}
    Qwen3.5-27B & \underline{0.1059} & \underline{0.2319} & \underline{0.0161} & \underline{0.4657} \\
    InternVL3.5-38B~\citep{internvl35} & 0.1032 & 0.1004 & 0.0003 & 0.4072 \\
    \cmidrule{1-5}
    \multicolumn{5}{l}{\textit{Open-source VLMs (medical)}} \\
    \cmidrule{1-5}
    Hulu-Med-32B~\citep{hulumed} & \underline{0.3476} & 0.1662 & 0.1256 & \underline{0.4687} \\
    Lingshu-32B~\citep{lingshu} & - & - & - & 0.2483 \\
    MedGemma 1.5 (4B)~\citep{medgemma15} & 0.1544 & 0.0910 & 0.0110 & 0.3394 \\
    RadFM~\citep{radfm} & 0.2472 & 0.1271 & \underline{0.2499} & 0.1632 \\
    CT-CHAT (70B)~\citep{ctrate} & 0.0000 & \underline{0.3408} & 0.1719 & - \\
    \bottomrule
  \end{tabular}
\end{table}

As shown in Table~\ref{tab:overall}, proprietary VLMs consistently outperform open-source counterparts across most datasets, even when provided with only three sampled slices as input. In particular, GPT-5.4 achieves the strongest overall performance. Among open-source general VLMs, Qwen3.5 shows relatively competitive results, while Hulu-Med~\citep{hulumed} performs best among medical VLMs. CT-CHAT~\citep{ctrate}, which is specifically trained for chest CT, demonstrates strong performance on the chest CT dataset CT-RATE~\citep{ctrate}, as well as on AMOS-MM~\citep{amos}, which partially includes thoracic regions.

Model performance on MIMIC-CXR~\citep{mimiccxr} is generally higher than on 3D datasets. This is likely because 2D chest X-ray data are more prevalent in training corpora, whereas 3D CT data remain underrepresented, leaving substantial room for improvement-particularly for open-source VLMs. Among 3D datasets, AMOS-MM~\citep{amos} yields the lowest performance, which can be attributed to its broader anatomical coverage, including abdominal, thoracic, and pelvic regions, resulting in increased complexity and variability. As 3D imaging continues to play an increasingly important role in clinical practice, advancing 3D medical image understanding is both a key research direction and a significant challenge for future VLMs.

\subsection{VLMs struggle with fine-grained attribute understanding}

\begin{figure}
  \centering
  \includegraphics[width=1\linewidth]{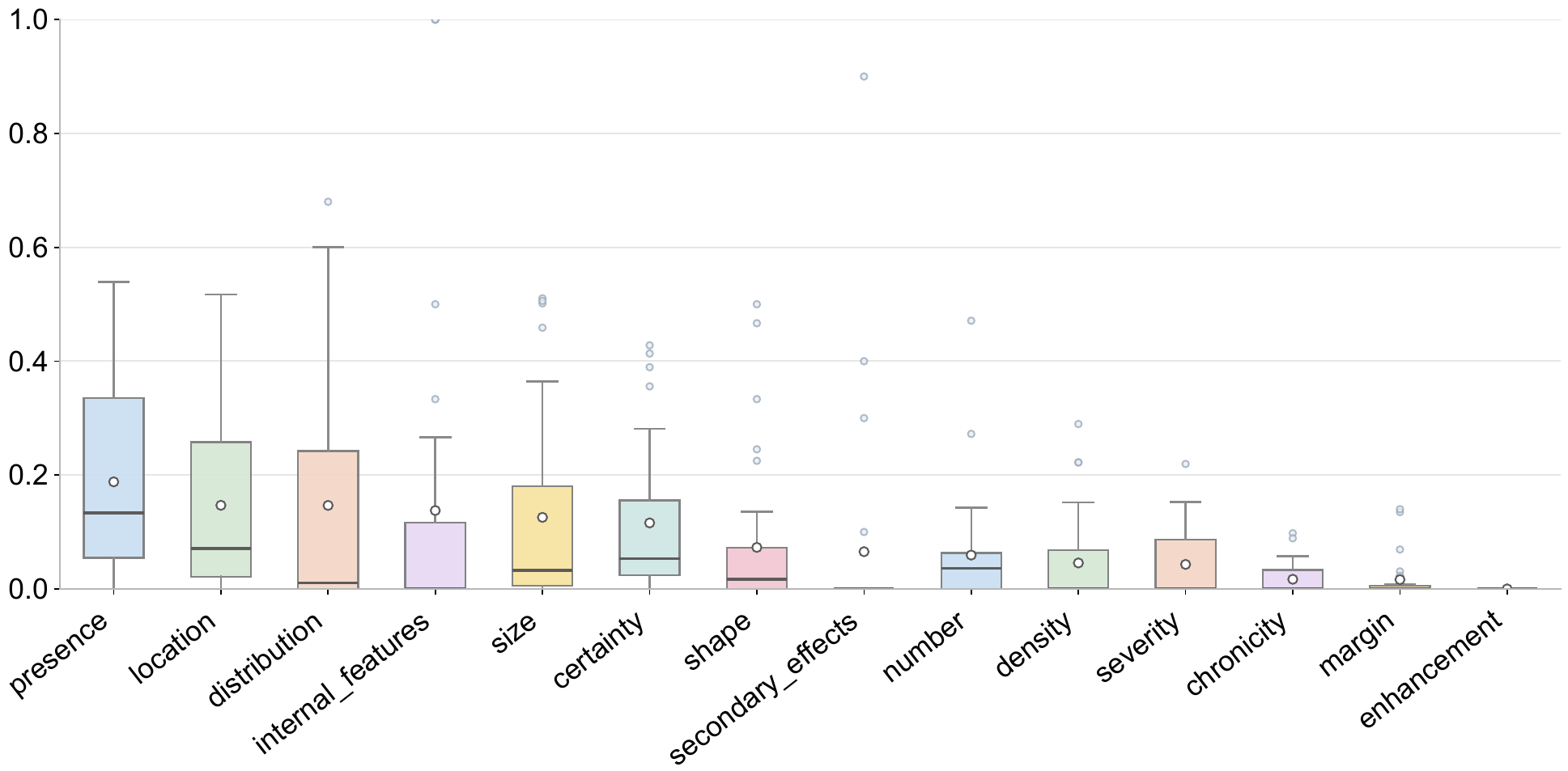}
  \caption{Attribute-level accuracy distribution across all dataset-model pairs. Y-axis denotes Accuracy, and each box shows the score distribution for one attribute. Higher boxes indicate attributes that are relatively easier for VLMs, while boxes concentrated near zero indicate more challenging attributes.}
  \label{fig:attribute}
\end{figure}

We analyze the accuracy across different attributes in ReportQA. As shown in Figure~\ref{fig:attribute}, performance varies substantially across attributes. Overall, \textit{presence} is the easiest attribute, with both its median and overall distribution higher than those of other attributes. This suggests that most VLMs are relatively stable and reliable in determining whether a clinical entity exists. Attributes such as \textit{location}, \textit{certainty}, and size also achieve moderate performance, but exhibit larger variance, indicating that their effectiveness depends more on specific models or datasets and is less consistent than \textit{presence}.

In contrast, \textit{margin}, \textit{chronicity}, and \textit{enhancement} are significantly more challenging. Their distributions are concentrated near zero, with low medians, indicating that models rarely achieve high accuracy on these fine-grained attributes. Attributes such as \textit{distribution}, \textit{internal features}, and \textit{secondary effects} show occasional high outliers but maintain low medians, suggesting that strong performance is limited to specific cases rather than being broadly achievable.

Overall, these results reveal a clear hierarchy in attribute difficulty: models perform well on presence-level judgments but struggle with fine-grained attributes, highlighting a key limitation of current VLMs and an important direction for future research on medical image analysis.

\subsection{Report-based SFT provides limited gains in fine-grained entity perception}

\begin{figure}
  \centering
  \includegraphics[width=1\linewidth]{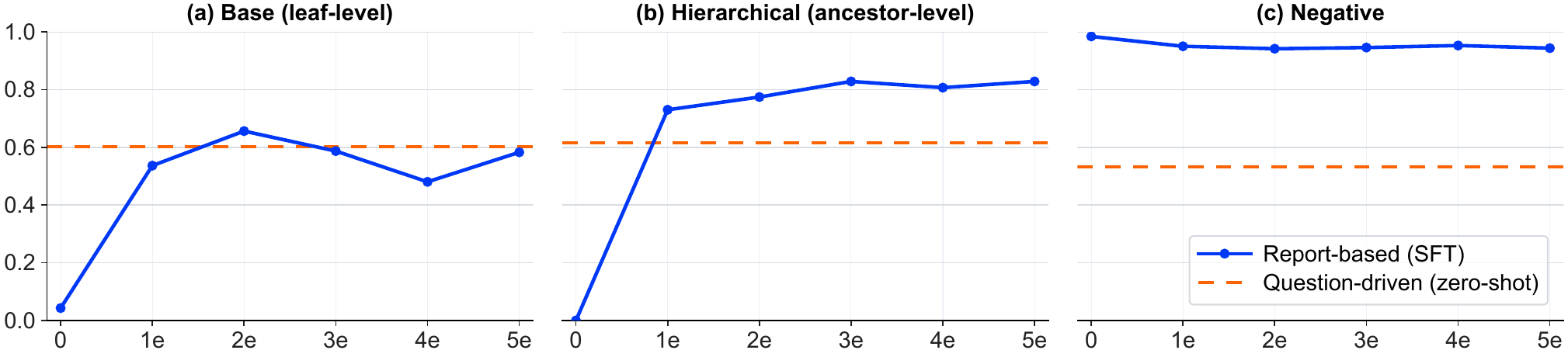}
  \caption{Performance of report-based SFT Qwen3.5-2B on CTRG-Brain for (a) Base (leaf-level), (b) Hierarchical (ancestor-level), and (c) Negative questions. On \textit{base (leaf-level)} questions targeting fine-grained entities, the \textbf{untrained} question-driven model performs comparably to SFT models.}
  \label{fig:epoch}
\end{figure}

Most existing RRG systems adopt a report-based inference paradigm, where the model generates a complete report in a single pass, during both zero-shot inference and supervised fine-tuning (SFT). In contrast, we explore a question-driven inference paradigm, where the same report is decomposed into multiple questions and the VLM is queried in a VQA-style manner. As shown in Figure~\ref{fig:epoch}, report-based SFT models outperform the \textbf{untrained} model under question-driven inference on \textit{hierarchical (ancestor-level)} questions, but achieve \textbf{comparable} performance on \textit{base (leaf-level)} questions.

As discussed in Section \ref{sec:qageneration}, \textit{base (leaf-level)} questions target relatively fine-grained clinical entities, whereas \textit{hierarchical (ancestor-level)} questions correspond to coarser-grained concepts. These results suggest that the gains from report-based SFT may primarily arise from improved report-style generation and coarse-grained structured reasoning, while improvements in fine-grained entity recognition remain limited and unstable. Therefore, current report-based training paradigms may encourage models to generate reports that better match the evaluation distribution, rather than consistently acquiring stronger fine-grained visual understanding capabilities.

\subsection{Report-based inference exhibits strong negative prior bias}

\begin{figure}
  \centering
  \includegraphics[width=1\linewidth]{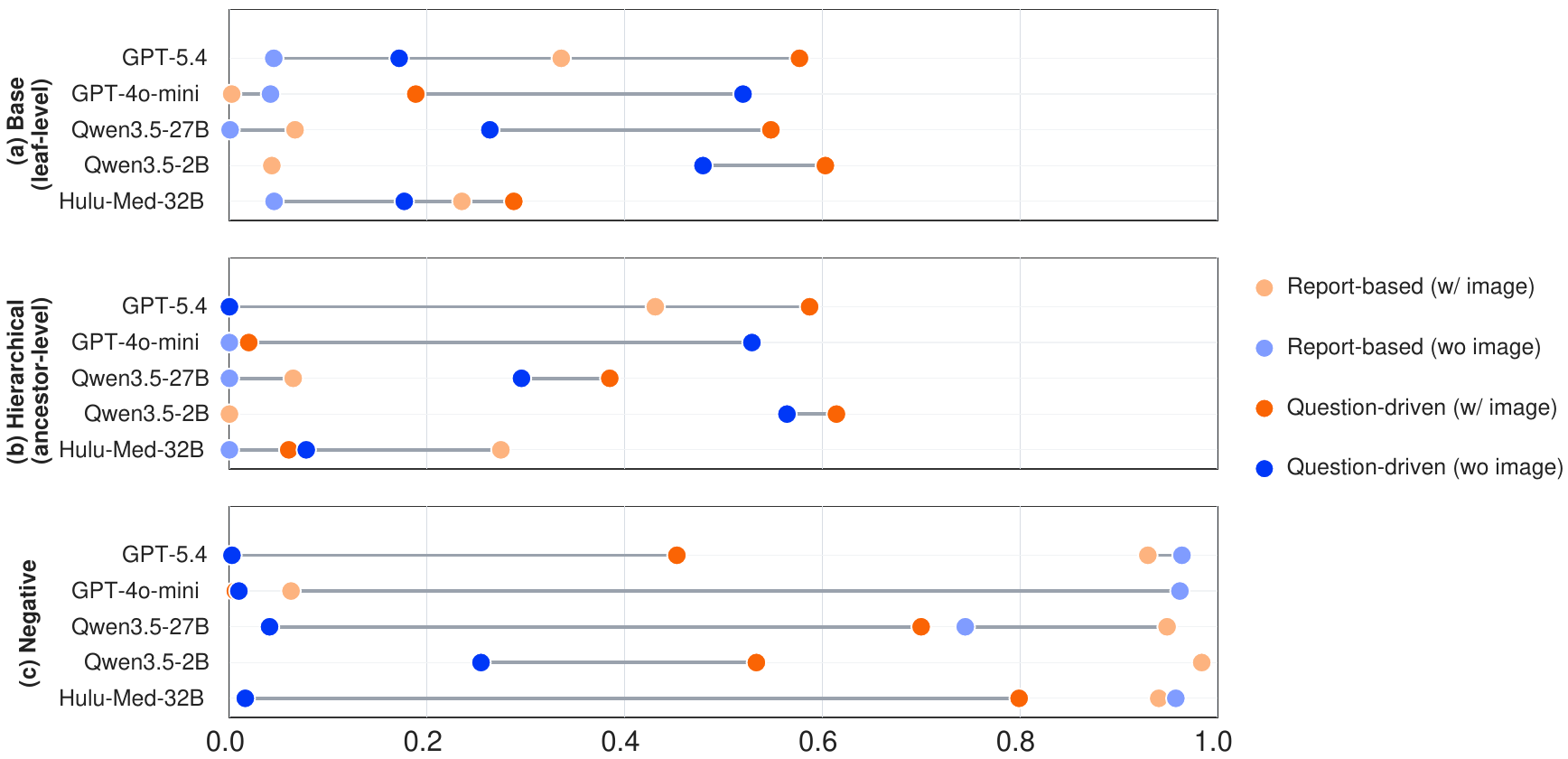}
  \caption{Impact of image input on performance under different inference paradigms.}
  \label{fig:zero}
\end{figure}

As shown in Figure~\ref{fig:zero}, models consistently exhibit a strong negative prior bias under the report-based inference paradigm. Even without image input, report-based inference tends to classify most clinical entities as negative. In contrast, question-driven inference shows a milder positive prior bias.

The difference in bias direction is likely related to training data distributions. Report-based inference typically learns the global distribution of complete radiology reports, where negative or routine descriptions are highly prevalent. As a result, when image evidence is absent, models tend to fall back to generic negative templates. In contrast, question-driven inference is centered around specific findings, lesions, or abnormalities. The questions themselves often contain stronger positive cues, which may encourage models to predict the presence of related abnormalities.

Overall, different inference paradigms activate different forms of textual priors. These biases are likely jointly influenced by both training data distributions and evaluation task design. Compared with report-based inference, \textbf{question-driven inference} may represent a more promising solution, as its bias can potentially be mitigated through the incorporation of negative training examples.

Additionally, the comparison between image input (\textit{w/ image}) and image-free input (\textit{wo image}) reveals varying degrees of \textit{mirage reasoning} across VLMs. Even without image input, VLMs can still produce seemingly plausible answers based solely on textual priors, achieving nontrivial or even relatively high scores. This phenomenon is more pronounced in VLMs with weaker visual capabilities, and in some cases leads to performance inversion, where \textit{wo image} outperforms \textit{w/ image}. For example, GPT-4o-mini exhibits clear inversion under both report-based and question-driven inference, suggesting it relies more heavily on textual priors than on stable visual grounding.

\section{Conclusion}

We present \textbf{ReportQA}, a QA-based RRE framework which supports \textbf{detailed quantitative analysis} of RRG systems. Through a well-designed pipeline, we construct approximately 660K QA pairs, averaging nearly 100 questions per report. Based on QA accuracy, we introduce \textbf{QAScore} metric. On RadEvalX~\citep{radevalx}, QAScore shows stronger alignment with radiologist judgments than existing metrics.

We conduct a comprehensive study across different VLMs, including proprietary VLMs, open-source general and medical VLMs. Results show that current VLMs still struggle with 3D medical tasks and fine-grained clinical attributes. Existing report-based inference paradigms have limited ability to learn fine-grained representations and exhibit strong negative prior biases. In contrast, we believe that \textbf{question-driven inference} paradigms, whose biases can be mitigated through mixed positive and negative training samples, represent a more promising solution for RRG systems.

We release all knowledge trees, structured reports, QA pairs, and pipeline code for QA construction and evaluation to support reproducibility and extensibility. Our framework can be readily applied to new datasets. Researchers can leverage more powerful judge models to further improve QA quality.


{
\small
\bibliographystyle{unsrtnat}
\bibliography{main}
}

\newpage
\appendix

\section{Judge models with different parameter scales}

\begin{figure}
  \centering
  \includegraphics[width=1\linewidth]{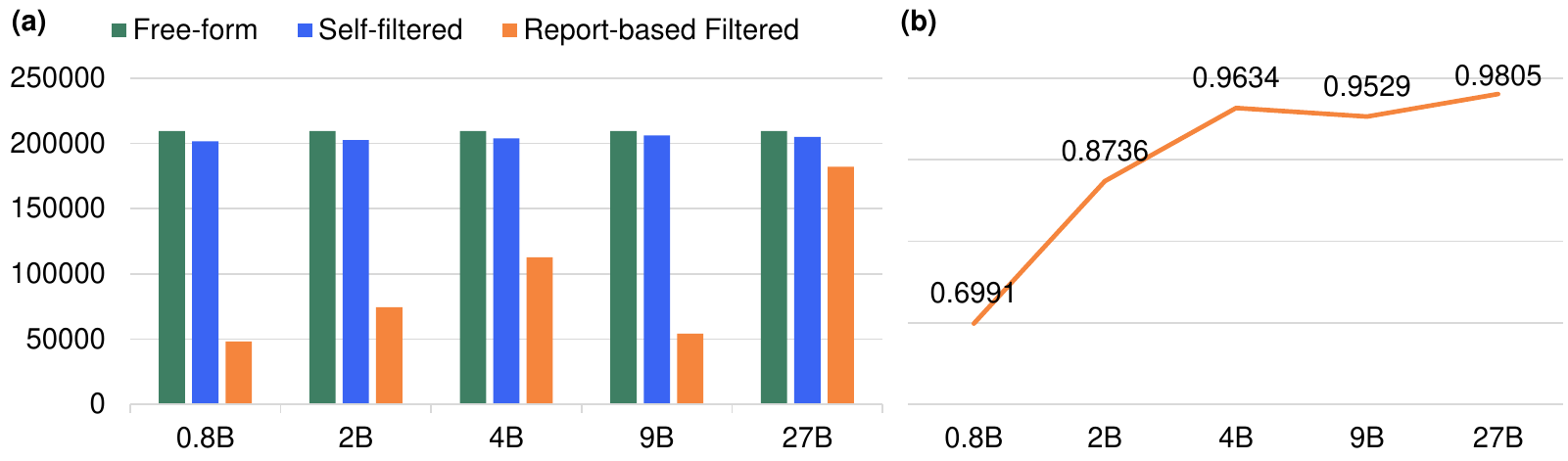}
  \caption{(a) Number of QA pairs retained after self-filtering and report-based filtering using judge models (Qwen3.5 family) with different parameter sizes. (b) QAScore based on ground-truth reports using judge models (Qwen3.5 family) with different parameter sizes.}
  \label{fig:judge}
\end{figure}

We analyze the impact of judge models with different parameter sizes (Qwen3.5 family) on both ReportQA construction, particularly QA filtering, and QA-based evaluation.

As shown in Figure~\ref{fig:judge} (a), different judge models lead to significantly different filtering ratios, with larger models such as 27B retaining substantially more QA pairs. Figure~\ref{fig:judge} (b) shows QAScores when ground-truth reports are used as context. The 0.8B model performs worst, achieving only around 0.7 accuracy, while the 27B model approaches near-perfect performance. 4B and 9B models also perform strongly, exceeding 0.95 accuracy, and remain feasible for deployment on consumer-grade GPUs.

These results reveal a trade-off between performance and efficiency. For optimal performance, both QA filtering and evaluation benefit from large-scale models such as 27B. However, during evaluation, smaller models such as 4B and 9B provide an alternative with minimal performance degradation.

\begin{figure}
  \centering
  \includegraphics[width=1\linewidth]{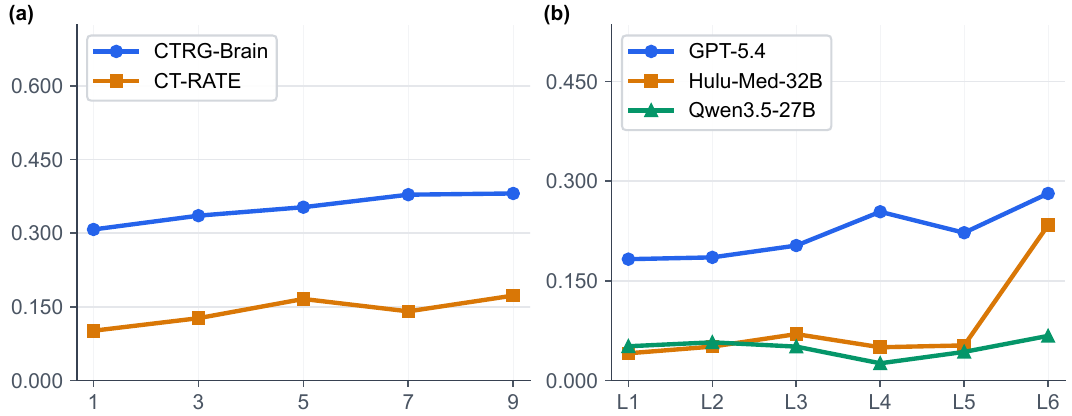}
  \caption{(a) Performance of GPT-5.4 with different numbers of input slices. (b) Performance of different models on CTRG-Brain~\citep{ctrg} under varying prompt specificity, where L1 denotes the least detailed prompts and L6 denotes the most detailed.}
  \label{fig:imageprompt}
\end{figure}

\section{Limited gains from increasing input slices}

As shown in Figure~\ref{fig:imageprompt} (a), model performance exhibits a modest upward trend as the number of input slices increases from 1 to 9, suggesting that additional slices provide more complete contextual information and yield incremental improvements. However, the overall gain remains limited, with a relatively flat curve, indicating that simply increasing the number of input slices does not fundamentally improve performance in radiology report generation.

\section{Prompt specificity improves performance but model-dependent}

As shown in Figure~\ref{fig:imageprompt} (b), as prompt specificity increases from L1 to L6, both GPT-5.4 and Hulu-Med-32B~\citep{hulumed} exhibit an upward trend in performance, indicating that more detailed and structured prompts help models better understand task requirements and organize their responses. In contrast, Qwen3.5-27B shows relatively limited performance variation, suggesting that general VLMs may lack sufficient training in RRG, and thus cannot fully benefit from increased prompt specificity due to gaps in domain knowledge and task-specific capabilities.

\section{Scaling alone is insufficient for RRG}

\begin{figure}
  \centering
  \includegraphics[width=1\linewidth]{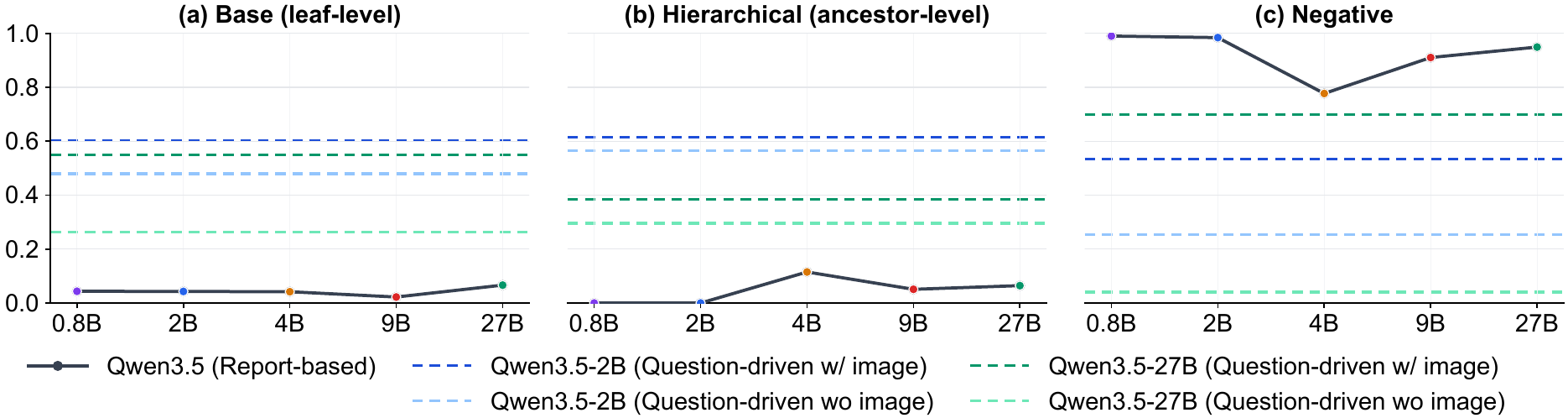}
  \caption{Performance of Qwen3.5 with different parameter sizes on CTRG-Brain~\citep{ctrg}.}
  \label{fig:para}
\end{figure}

As shown in Figure~\ref{fig:para}, under the report-based inference setting, Qwen3.5 family does not exhibit a consistent performance gain with increasing model size. Curves across different parameter scales remain at similar levels. This suggests that, for the complex and relatively unfamiliar task of RRG, scaling model size alone does not naturally translate into stronger task generalization, and model capacity is not the primary performance bottleneck.

Comparing question-driven inference (using images) to the report-based setting, further reveals that VLMs possess a certain level of clinical entity recognition ability, but this capability is not fully utilized in RRG. In particular, performance under question-driven setting is consistently higher, indicating that when the task is decomposed into explicit clinical queries, VLMs can more effectively leverage the input information. In contrast, report-based inference requires not only identifying clinical entities, but also organizing report structure, selecting appropriate expressions, and maintaining consistency with visual evidence. As a result, the bottleneck likely lies in aligning visual evidence with structured text generation, rather than in low-level recognition alone.

The fact that question-driven inference without images still achieves non-trivial performance suggests the presence of strong textual priors, where VLMs rely on statistical patterns in the questions even without visual input. However, question-driven inference with images consistently performs better, indicating that visual information still provides meaningful gains. Thus, the issue is not the absence of visual reasoning, but rather the incomplete utilization of visual signals. Furthermore, comparing smaller and larger models, such as 2B and 27B, shows a relative decline in \textit{wo image} performance, suggesting that reliance on textual priors may weaken with scale, although this trend requires further validation across more models.

\section{Examples of prompt and knowledge trees}
\label{sec:prompt}

\begin{figure}
  \centering
  \includegraphics[width=1\linewidth]{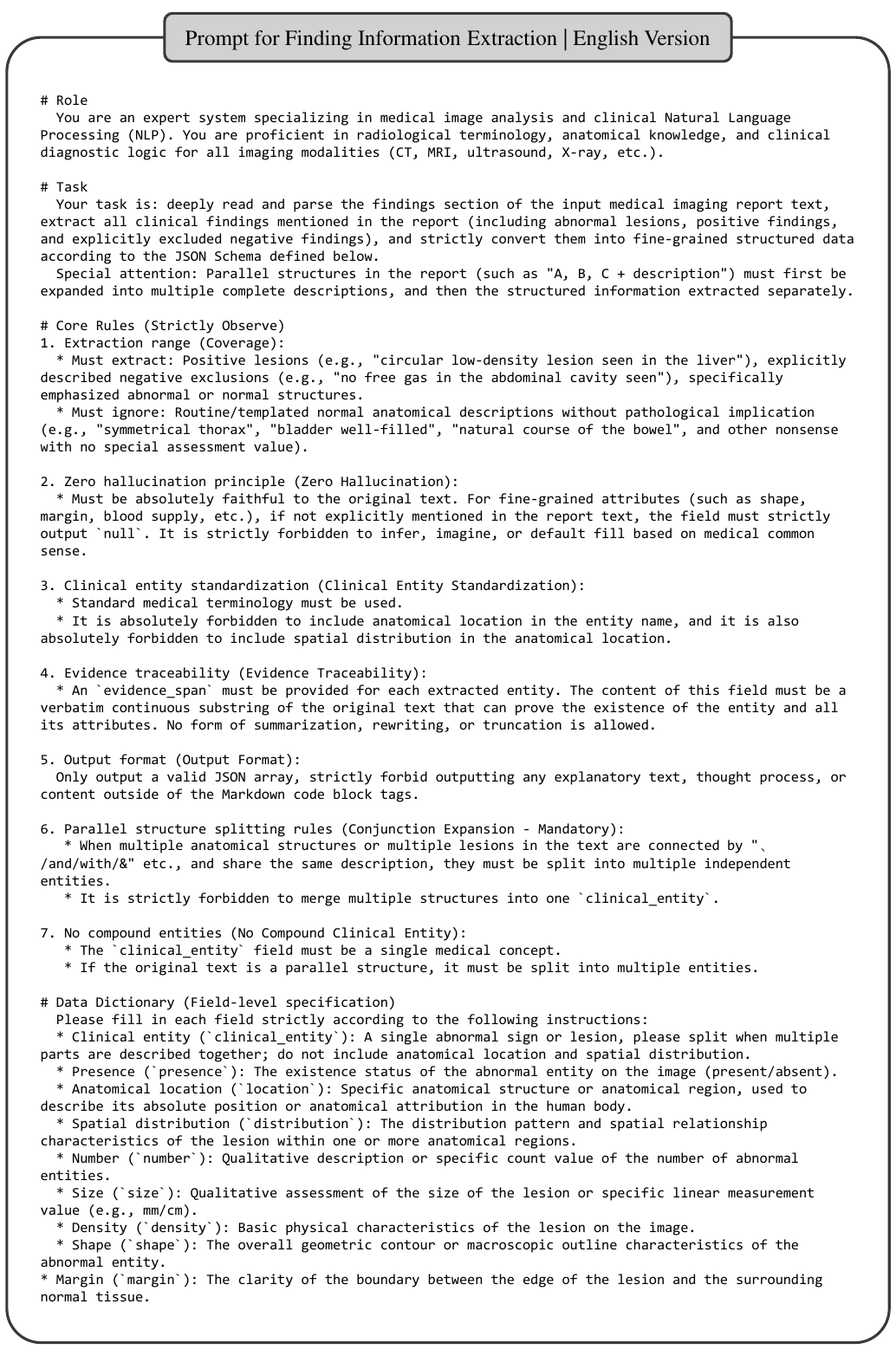}
  \caption{Prompt for finding information extraction (English version).}
  \label{fig:1}
\end{figure}

\begin{figure}
  \centering
  \includegraphics[width=1\linewidth]{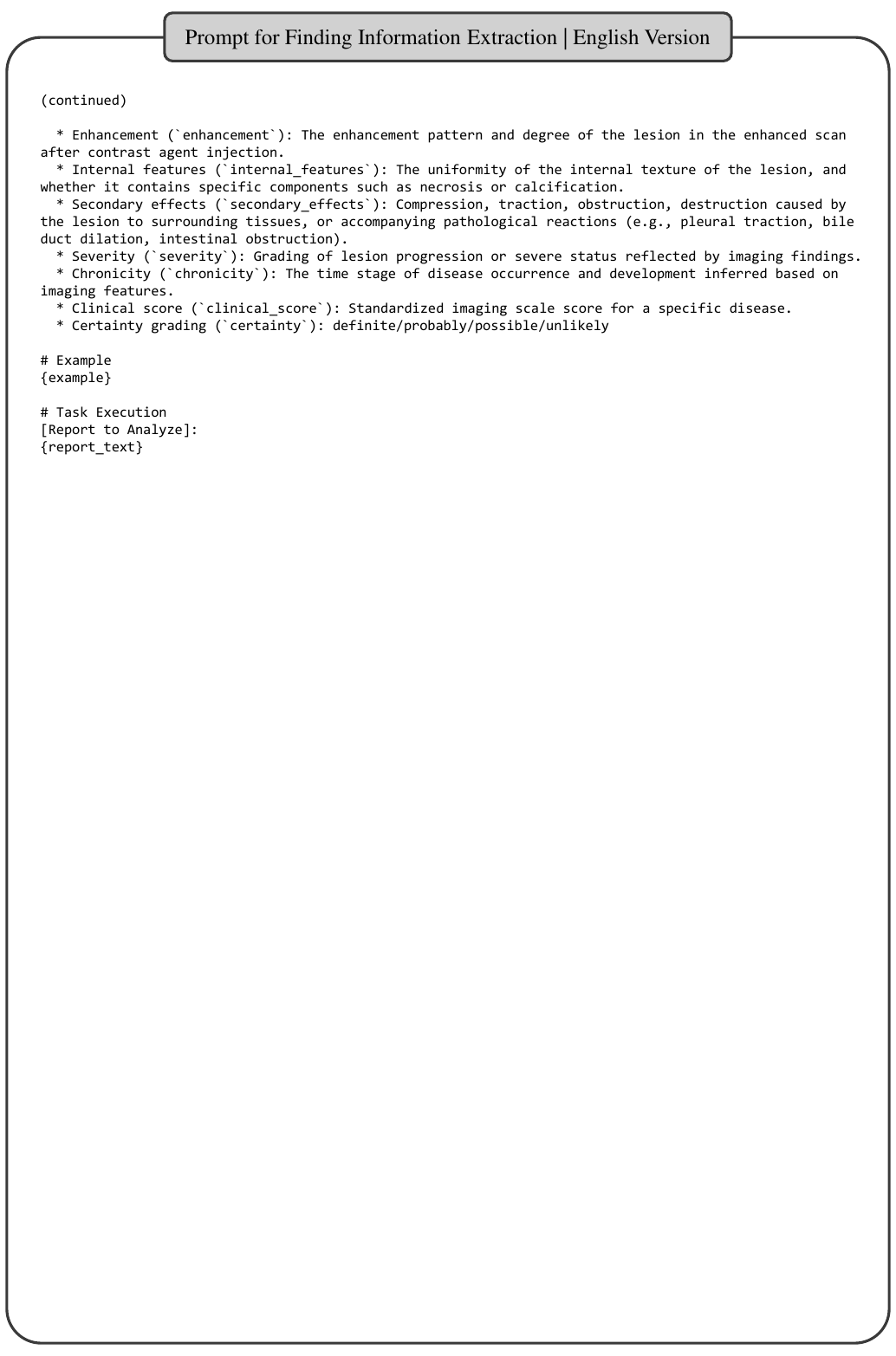}
  \caption{Prompt for finding information extraction (English version).}
  \label{fig:2}
\end{figure}

\begin{figure}
  \centering
  \includegraphics[width=1\linewidth]{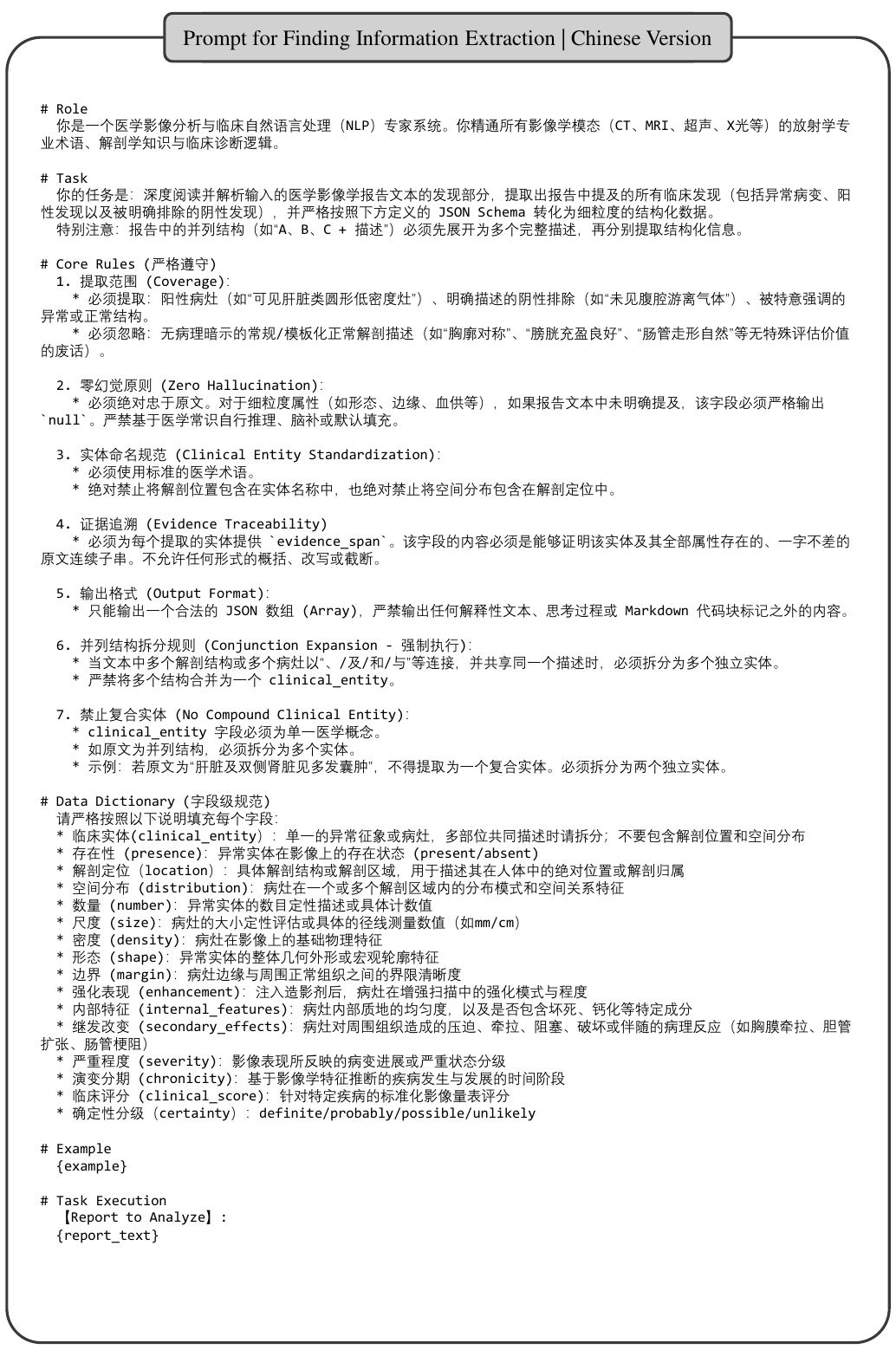}
  \caption{Prompt for finding information extraction (Chinese version).}
  \label{fig:3}
\end{figure}

\begin{figure}
  \centering
  \includegraphics[width=1\linewidth]{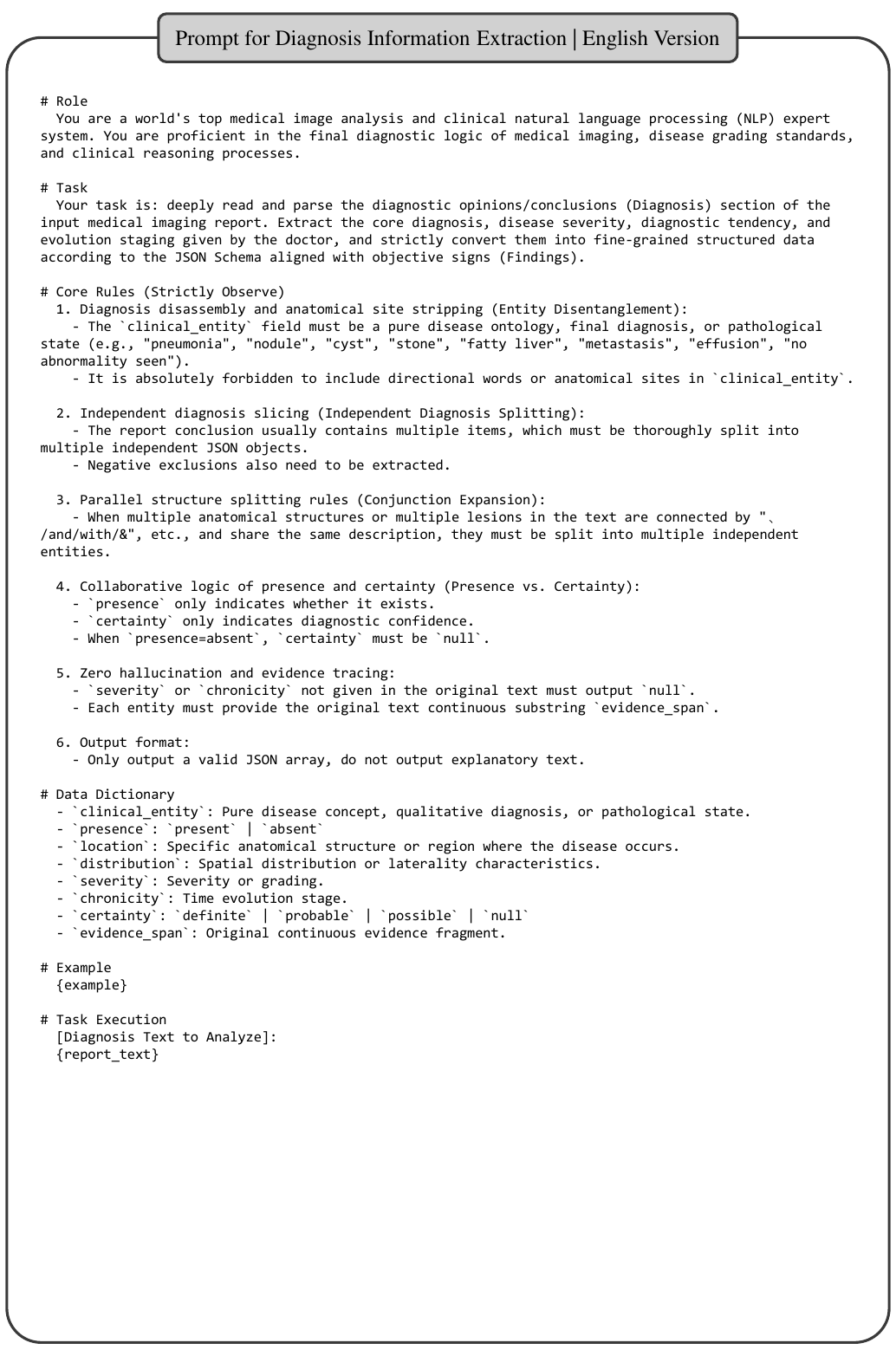}
  \caption{Prompt for diagnosis information extraction (English version).}
  \label{fig:4}
\end{figure}

\begin{figure}
  \centering
  \includegraphics[width=1\linewidth]{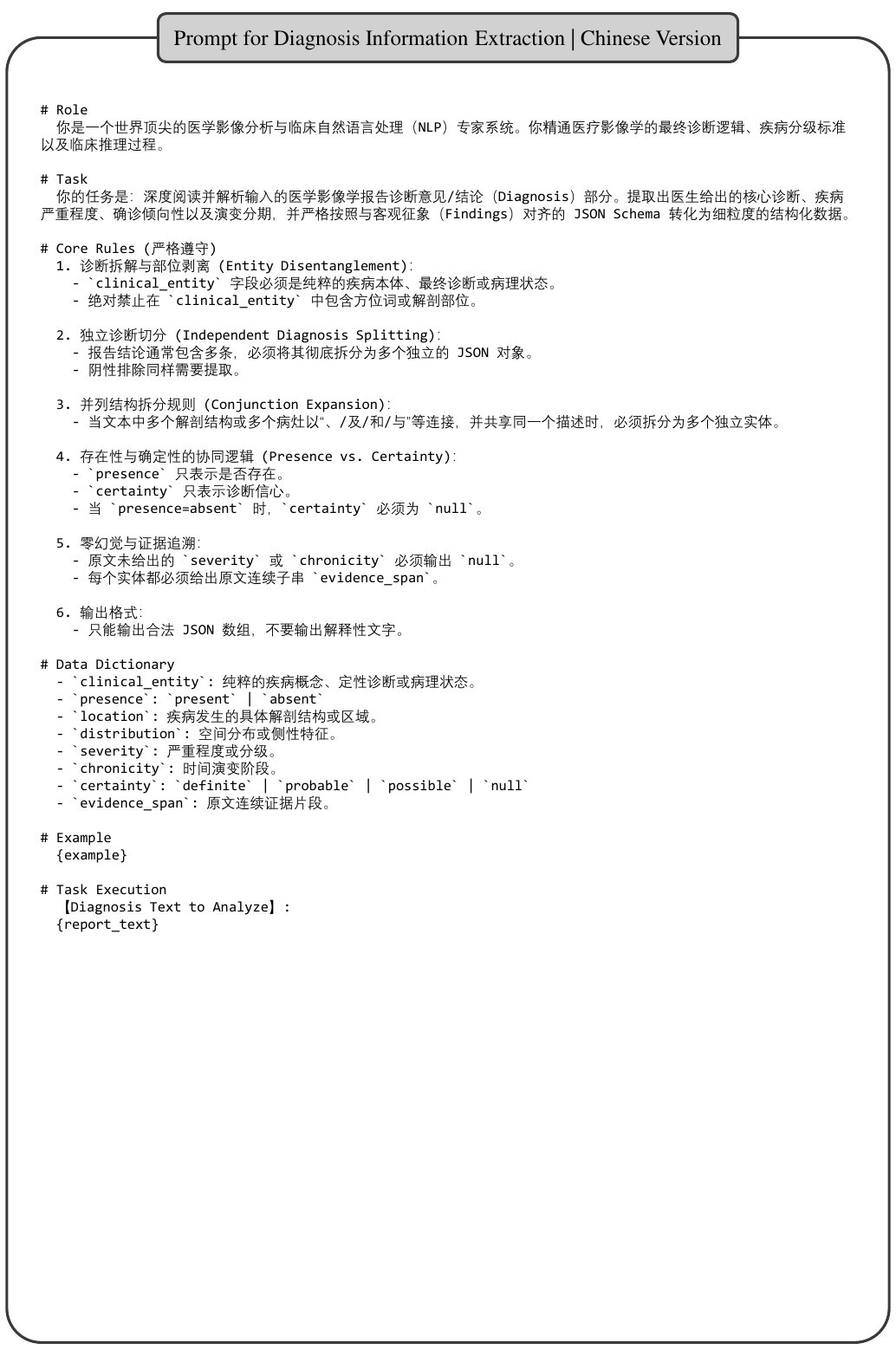}
  \caption{Prompt for diagnosis information extraction (Chinese version).}
  \label{fig:5}
\end{figure}

\begin{figure}
  \centering
  \includegraphics[width=1\linewidth]{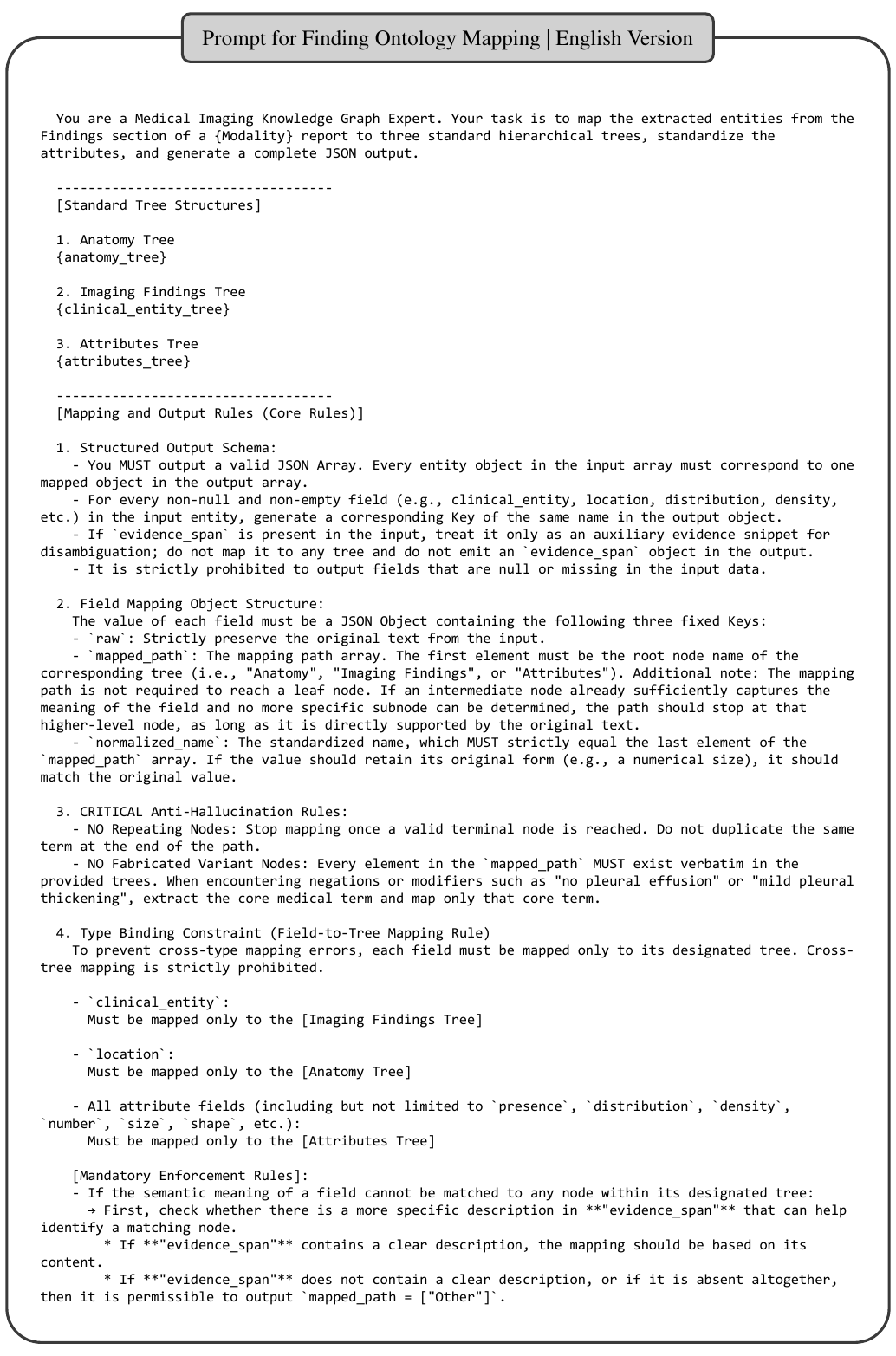}
  \caption{Prompt for finding ontology mapping (English version).}
  \label{fig:6}
\end{figure}

\begin{figure}
  \centering
  \includegraphics[width=1\linewidth]{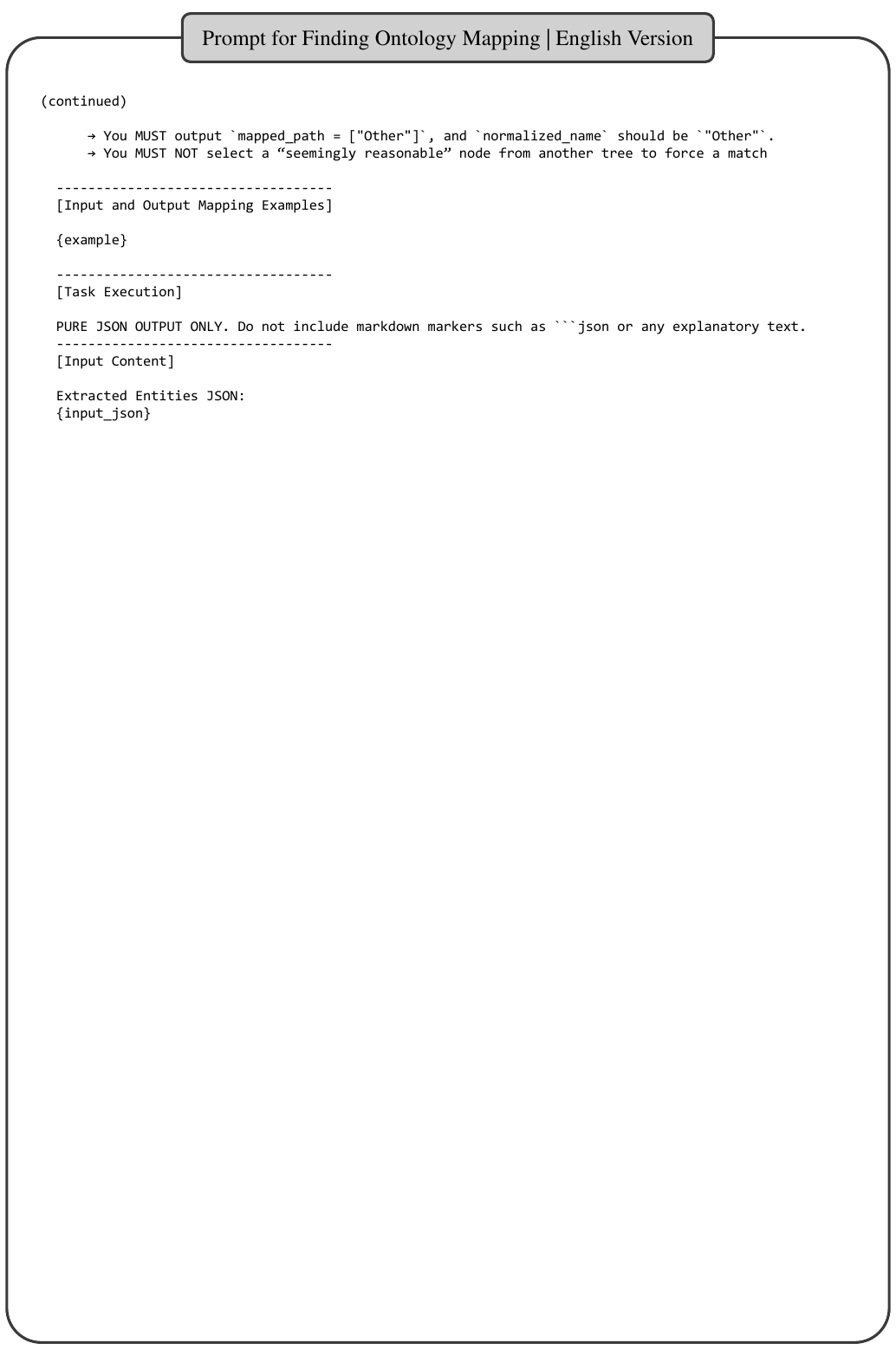}
  \caption{Prompt for finding ontology mapping (English version).}
  \label{fig:7}
\end{figure}

\begin{figure}
  \centering
  \includegraphics[width=1\linewidth]{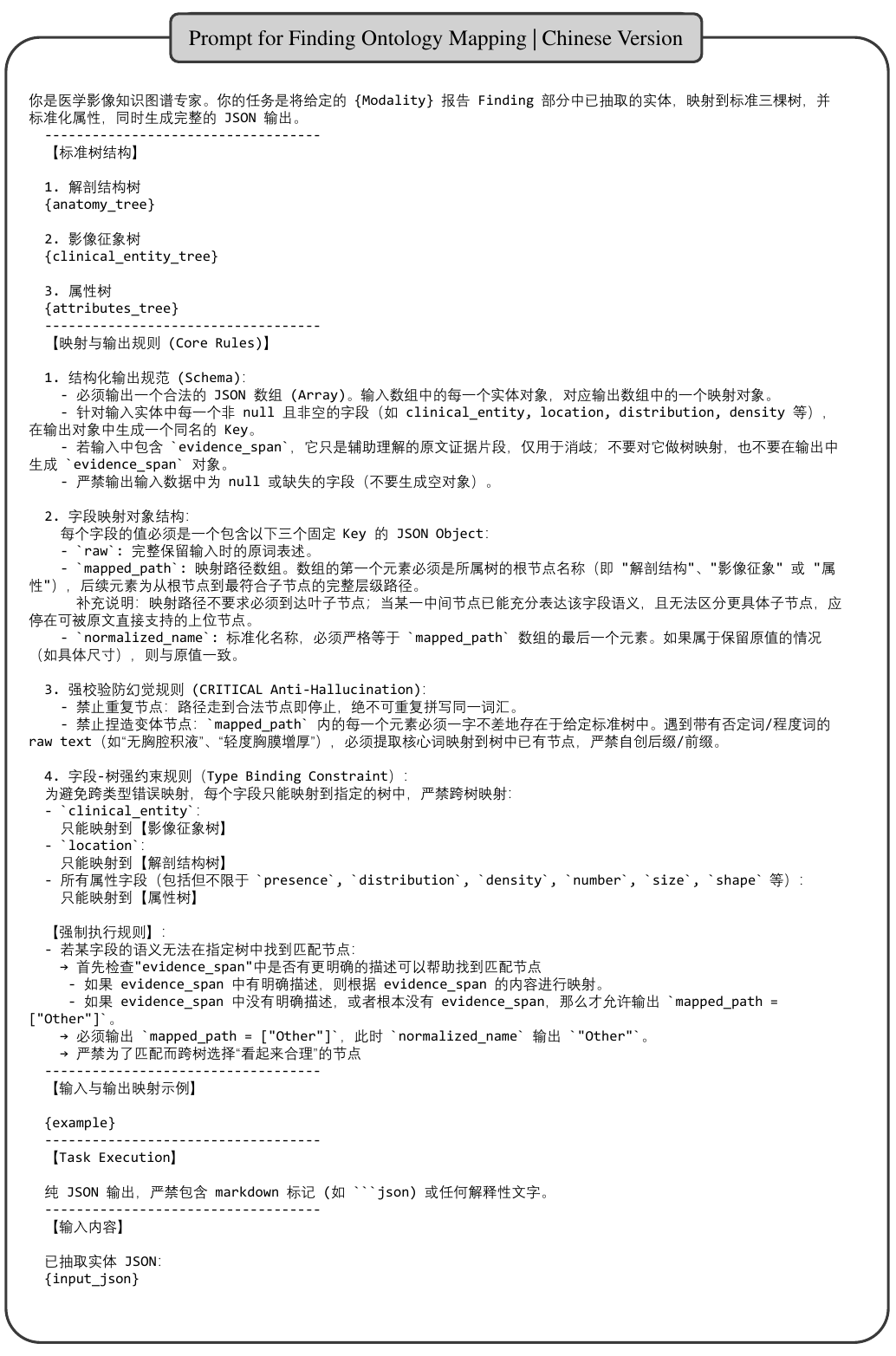}
  \caption{Prompt for finding ontology mapping (Chinese version).}
  \label{fig:8}
\end{figure}

\begin{figure}
  \centering
  \includegraphics[width=1\linewidth]{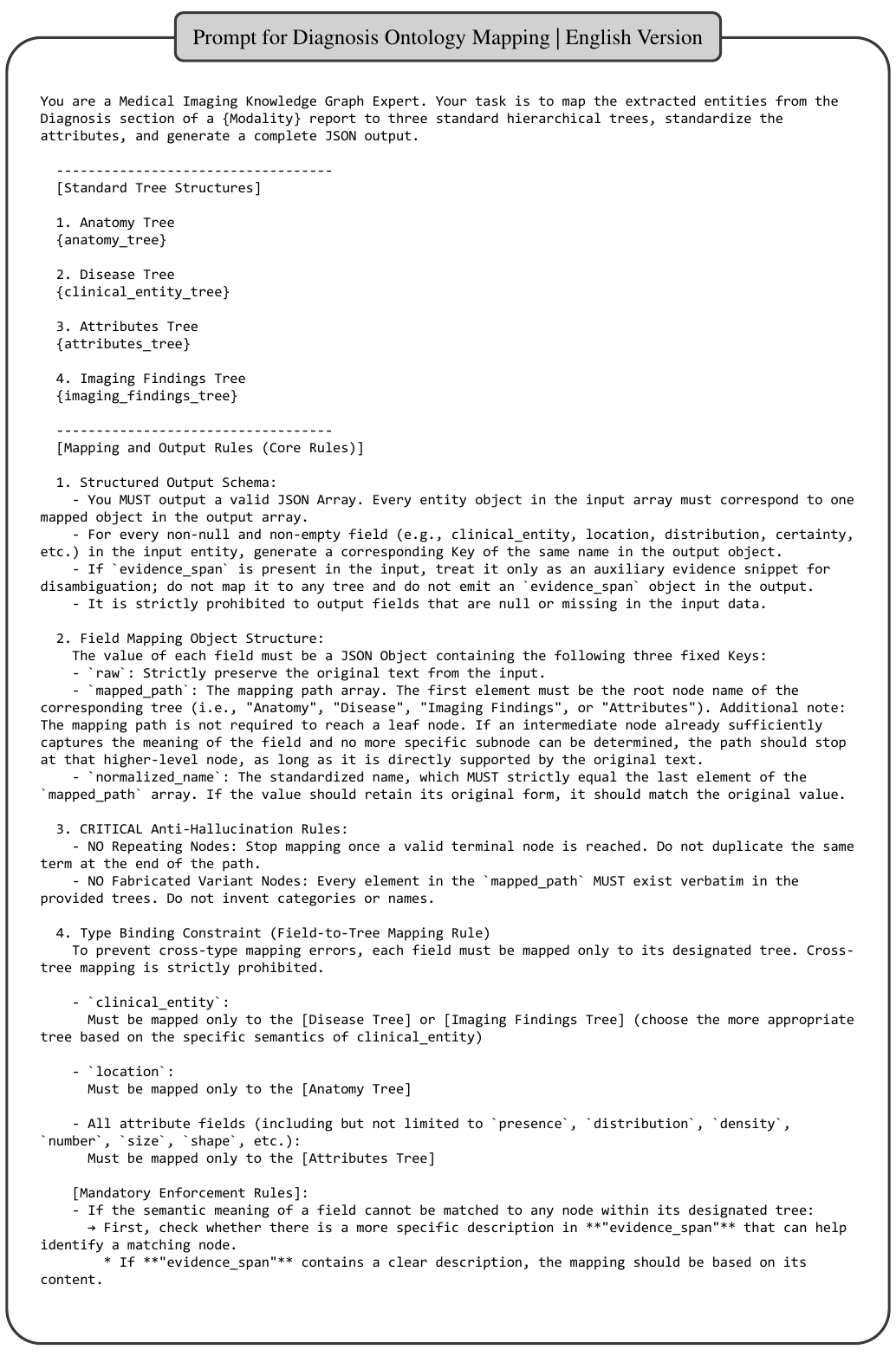}
  \caption{Prompt for diagnosis ontology mapping (English version).}
  \label{fig:9}
\end{figure}

\begin{figure}
  \centering
  \includegraphics[width=1\linewidth]{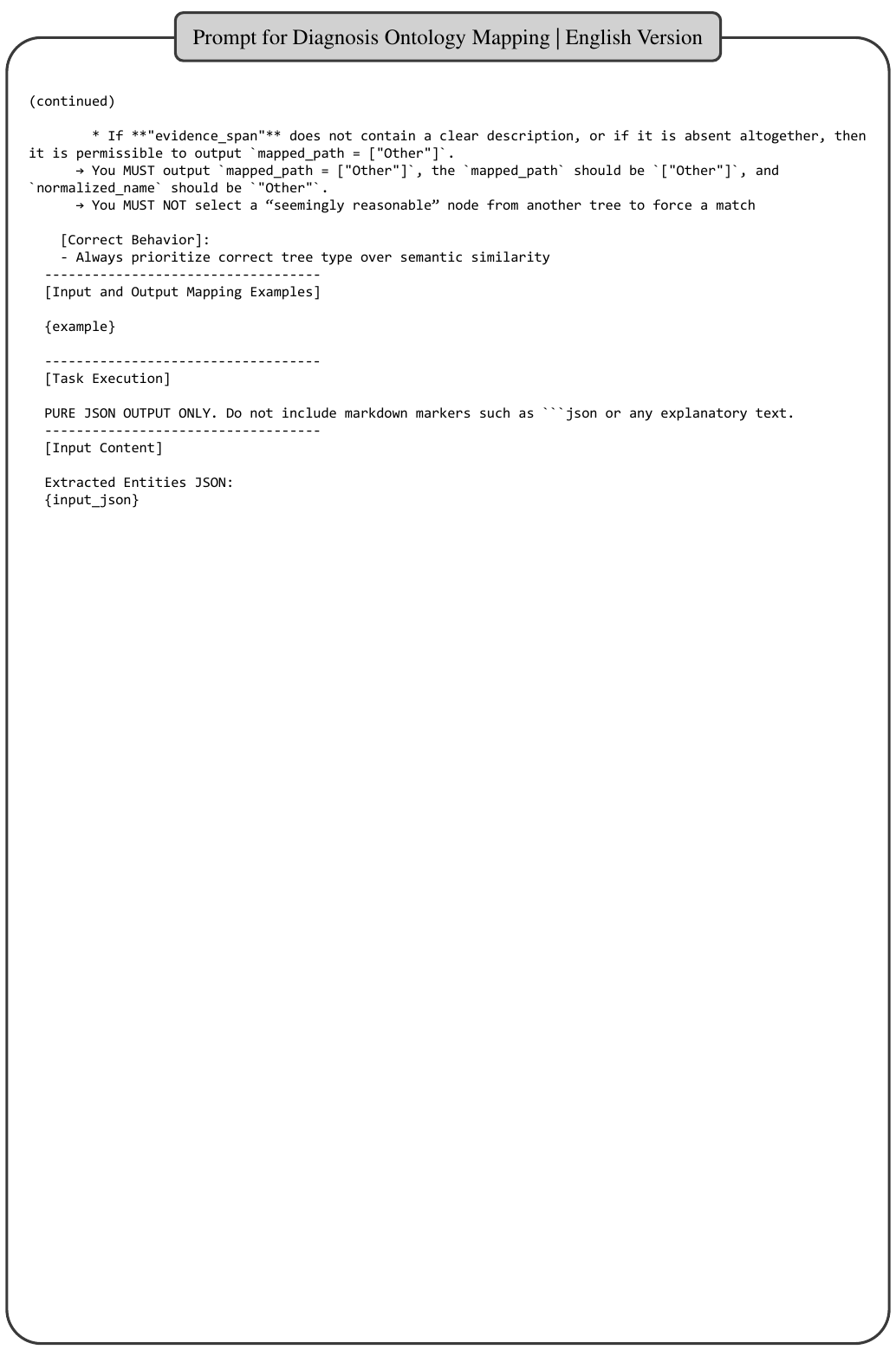}
  \caption{Prompt for diagnosis ontology mapping (English version).}
  \label{fig:10}
\end{figure}

\begin{figure}
  \centering
  \includegraphics[width=1\linewidth]{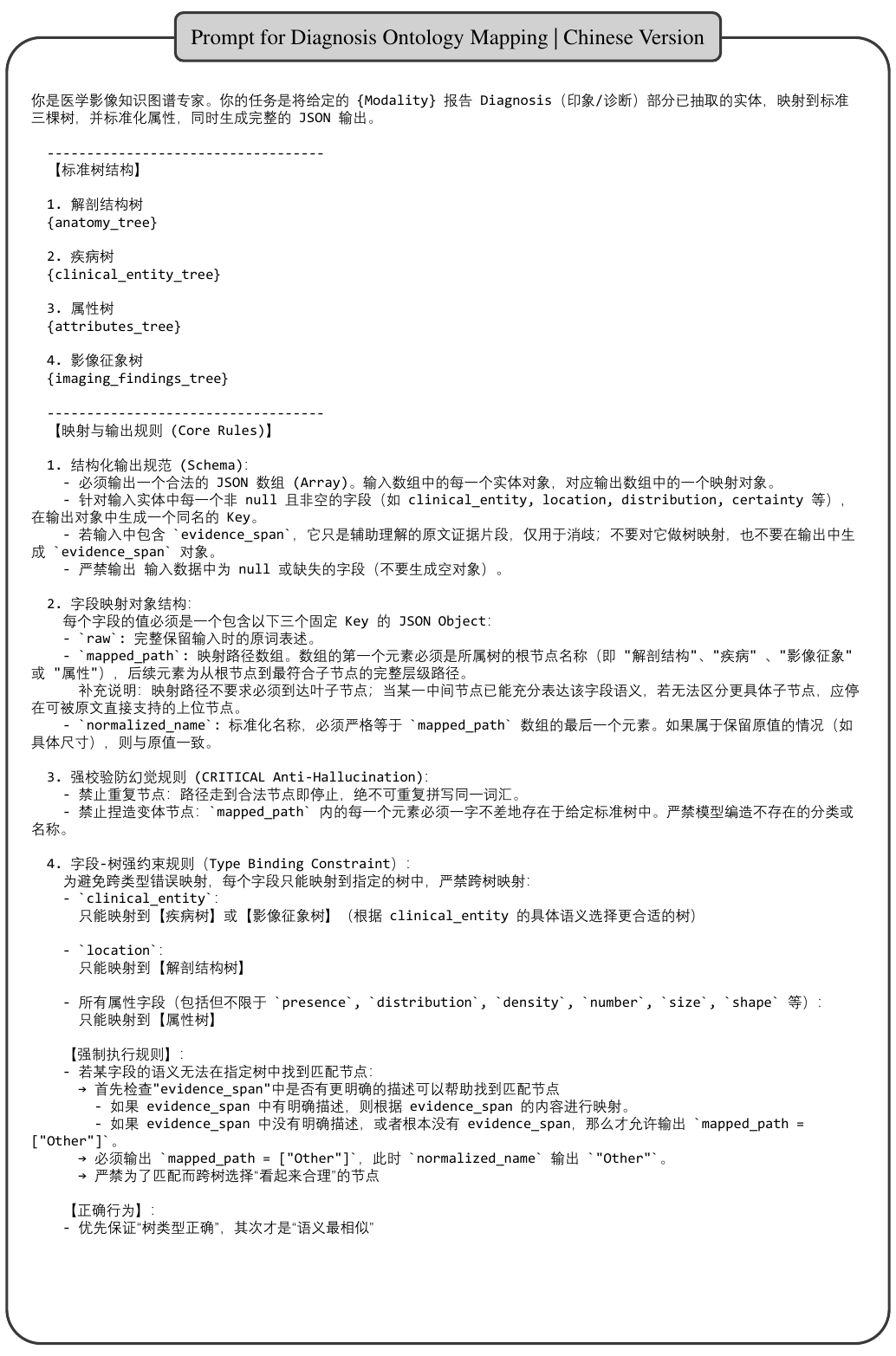}
  \caption{Prompt for diagnosis ontology mapping (Chinese version).}
  \label{fig:11}
\end{figure}

\begin{figure}
  \centering
  \includegraphics[width=1\linewidth]{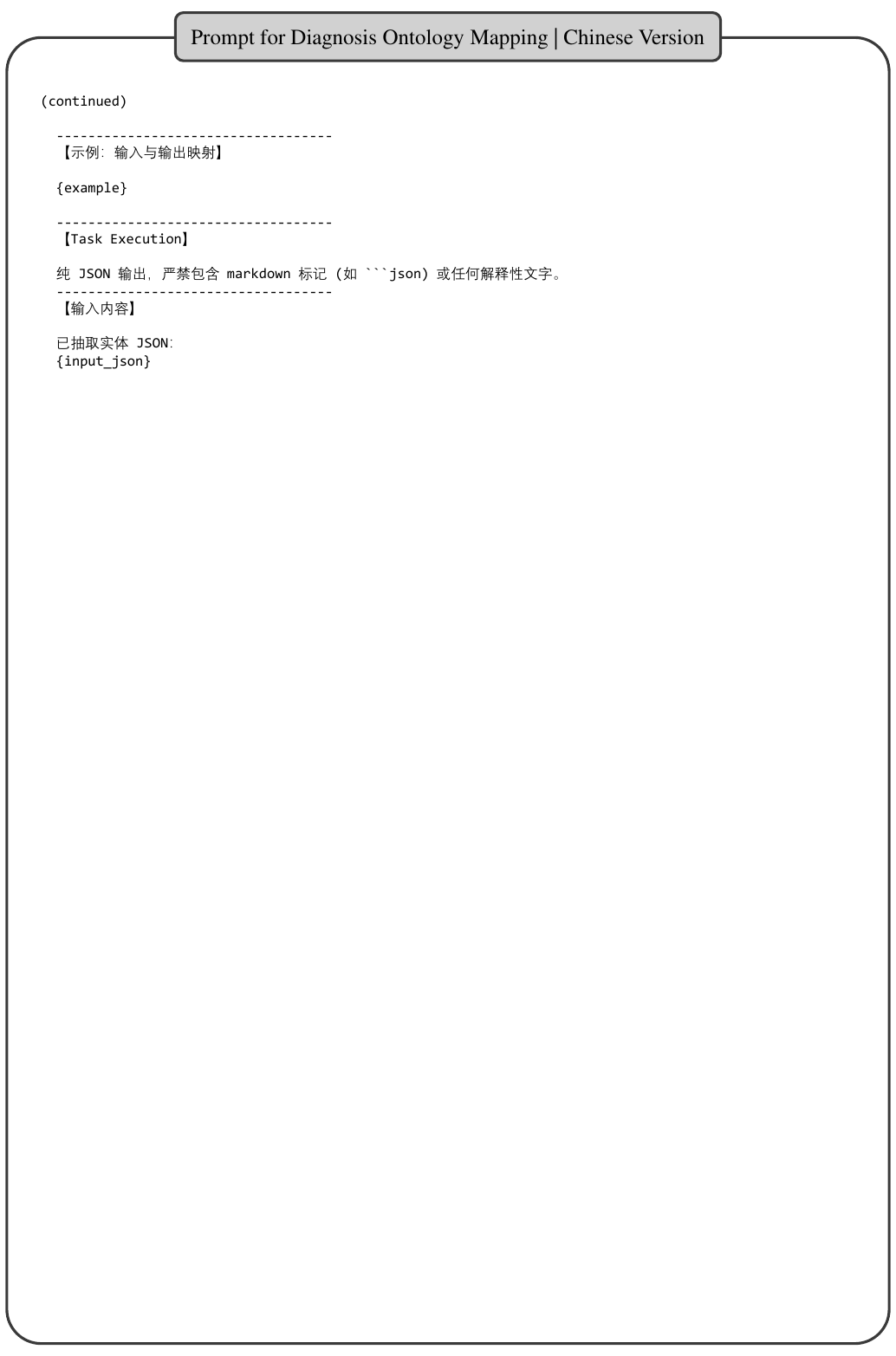}
  \caption{Prompt for diagnosis ontology mapping (Chinese version).}
  \label{fig:12}
\end{figure}

\begin{figure}
  \centering
  \includegraphics[width=1\linewidth]{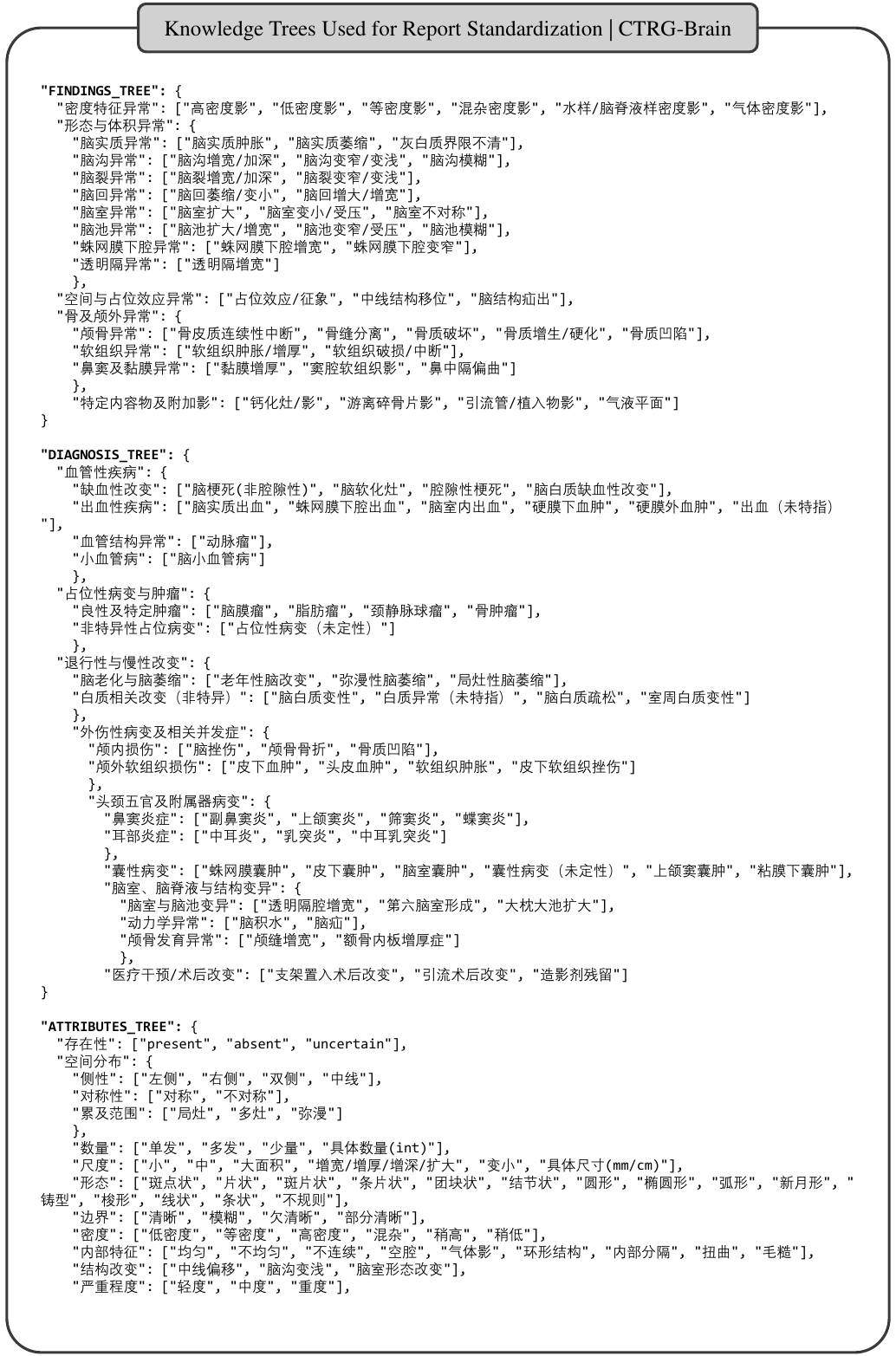}
  \caption{Knowledge trees for CTRG-Brain~\citep{ctrg}.}
  \label{fig:13}
\end{figure}

\begin{figure}
  \centering
  \includegraphics[width=1\linewidth]{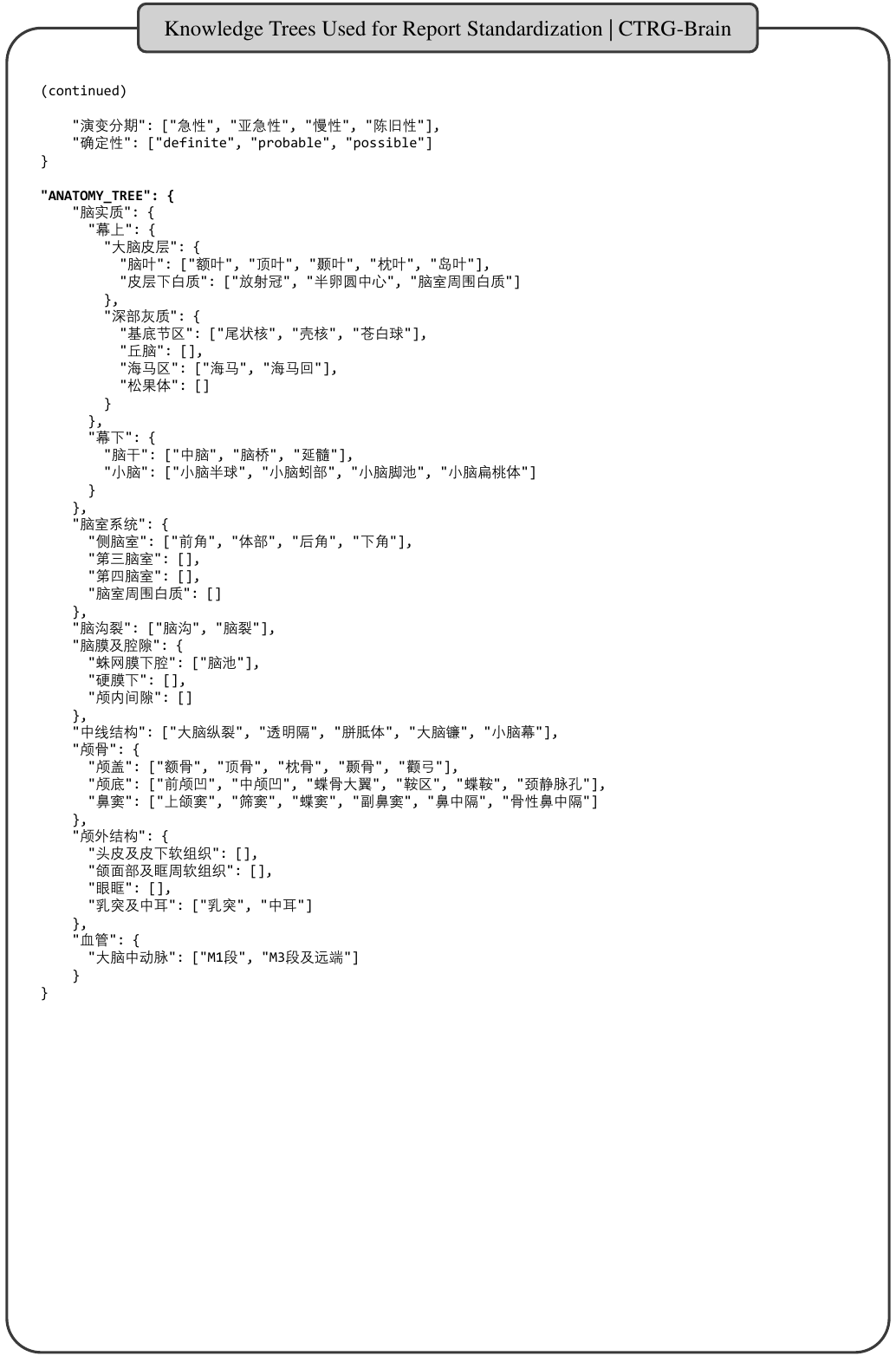}
  \caption{Knowledge trees for CTRG-Brain~\citep{ctrg}.}
  \label{fig:14}
\end{figure}

\begin{figure}
  \centering
  \includegraphics[width=1\linewidth]{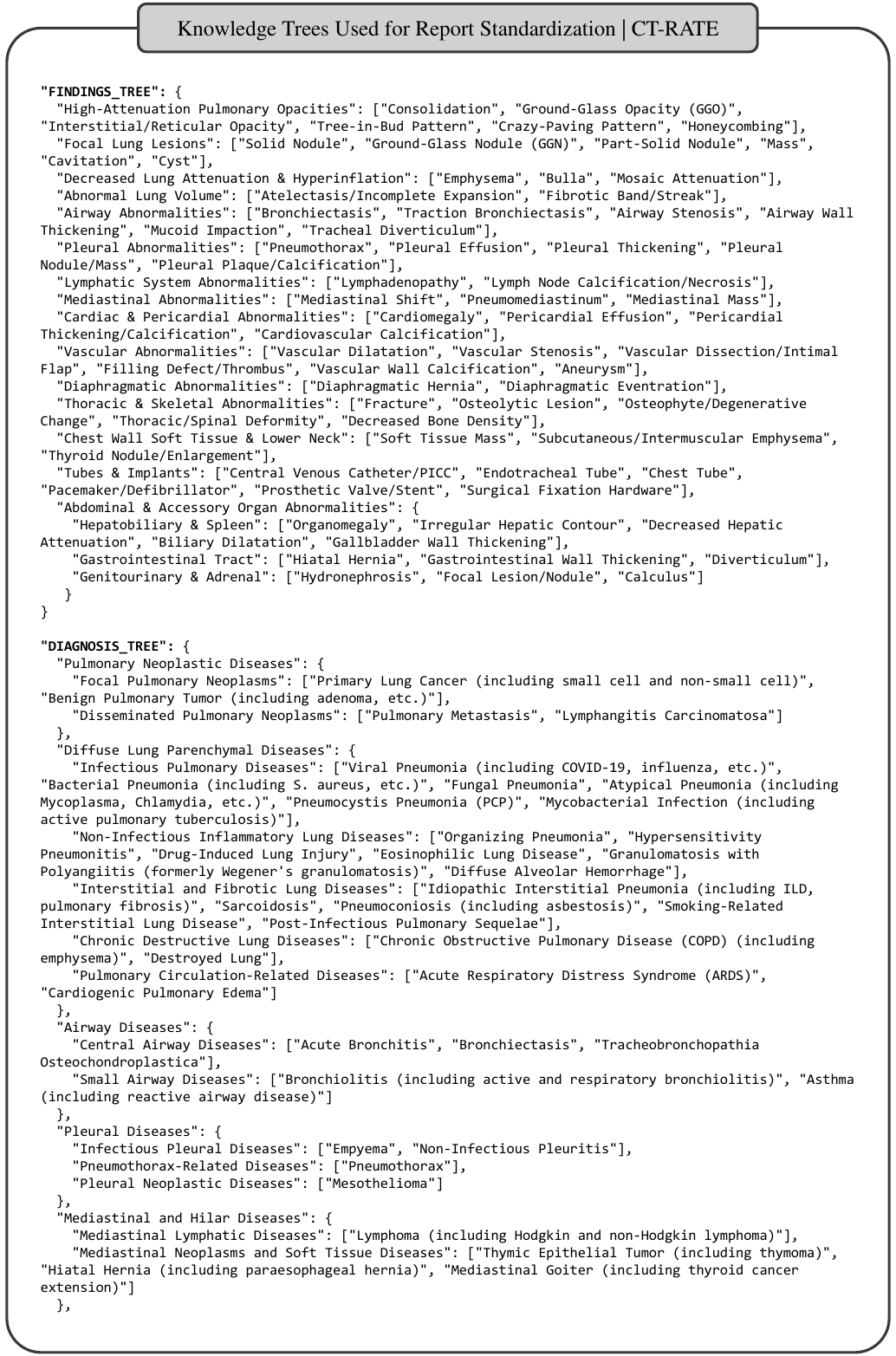}
  \caption{Knowledge trees for CT-RATE~\citep{ctrate}.}
  \label{fig:15}
\end{figure}

\begin{figure}
  \centering
  \includegraphics[width=1\linewidth]{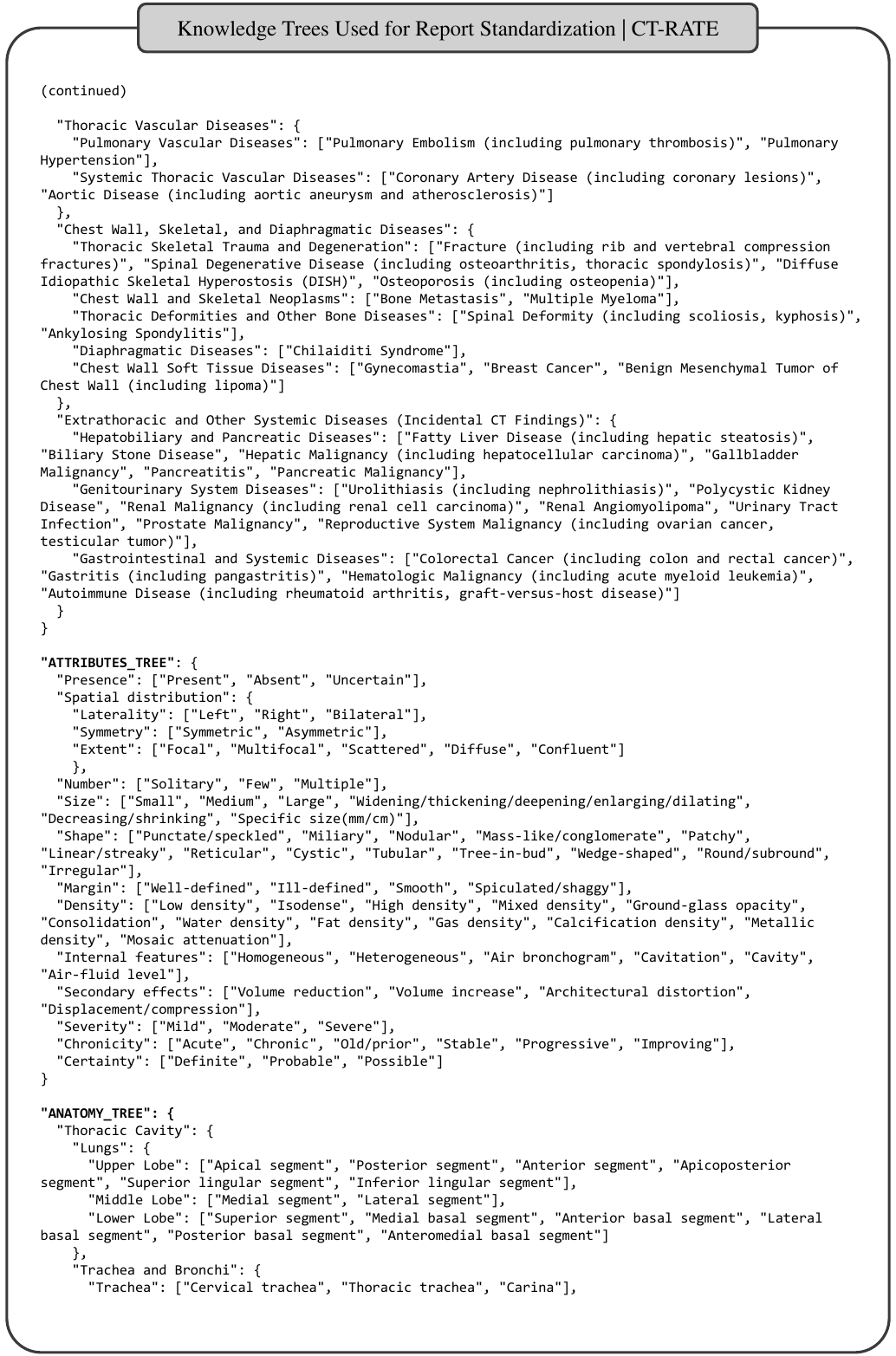}
  \caption{Knowledge trees for CT-RATE~\citep{ctrate}.}
  \label{fig:16}
\end{figure}

\begin{figure}
  \centering
  \includegraphics[width=1\linewidth]{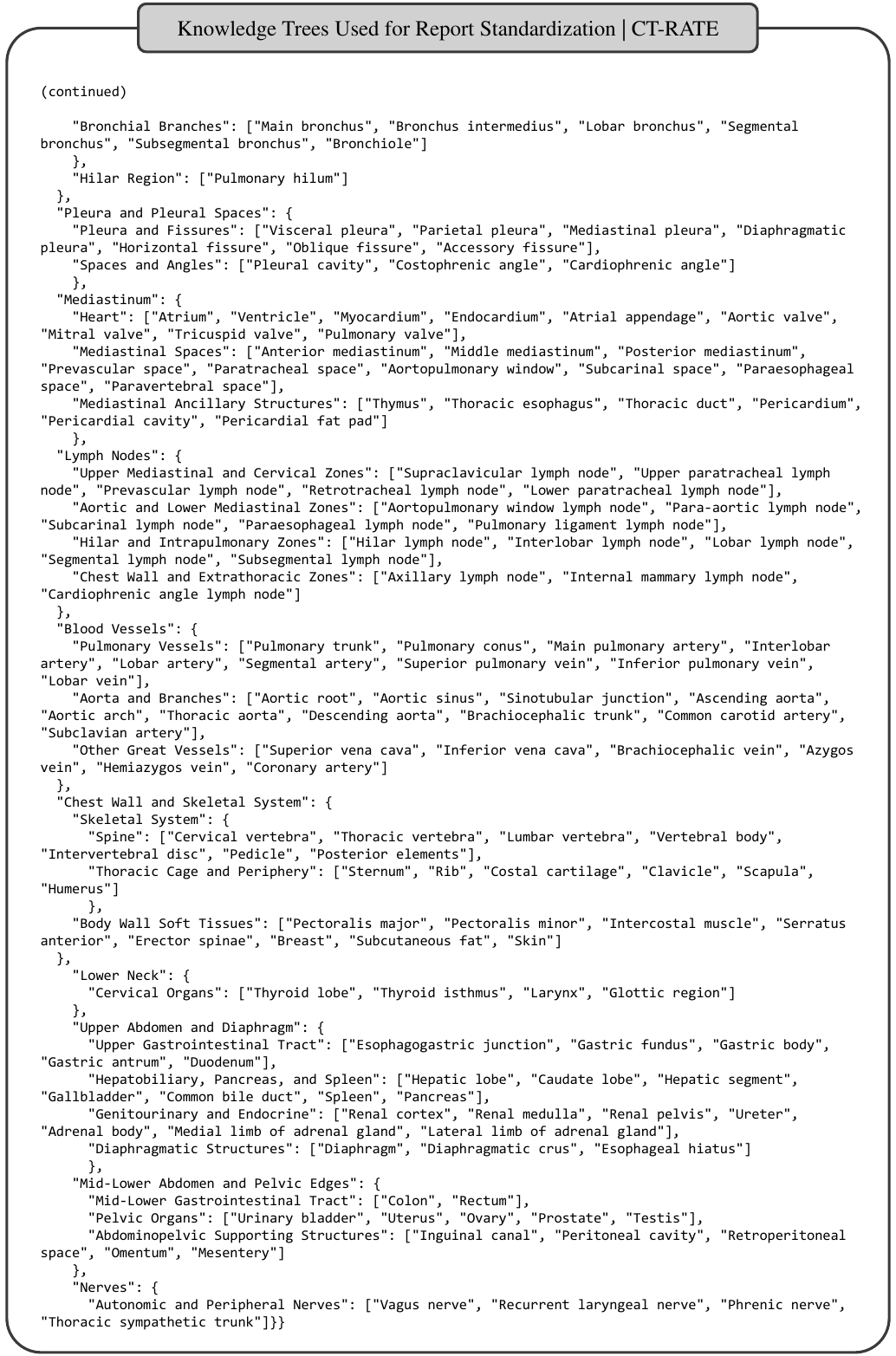}
  \caption{Knowledge trees for CT-RATE~\citep{ctrate}.}
  \label{fig:17}
\end{figure}

\begin{figure}
  \centering
  \includegraphics[width=1\linewidth]{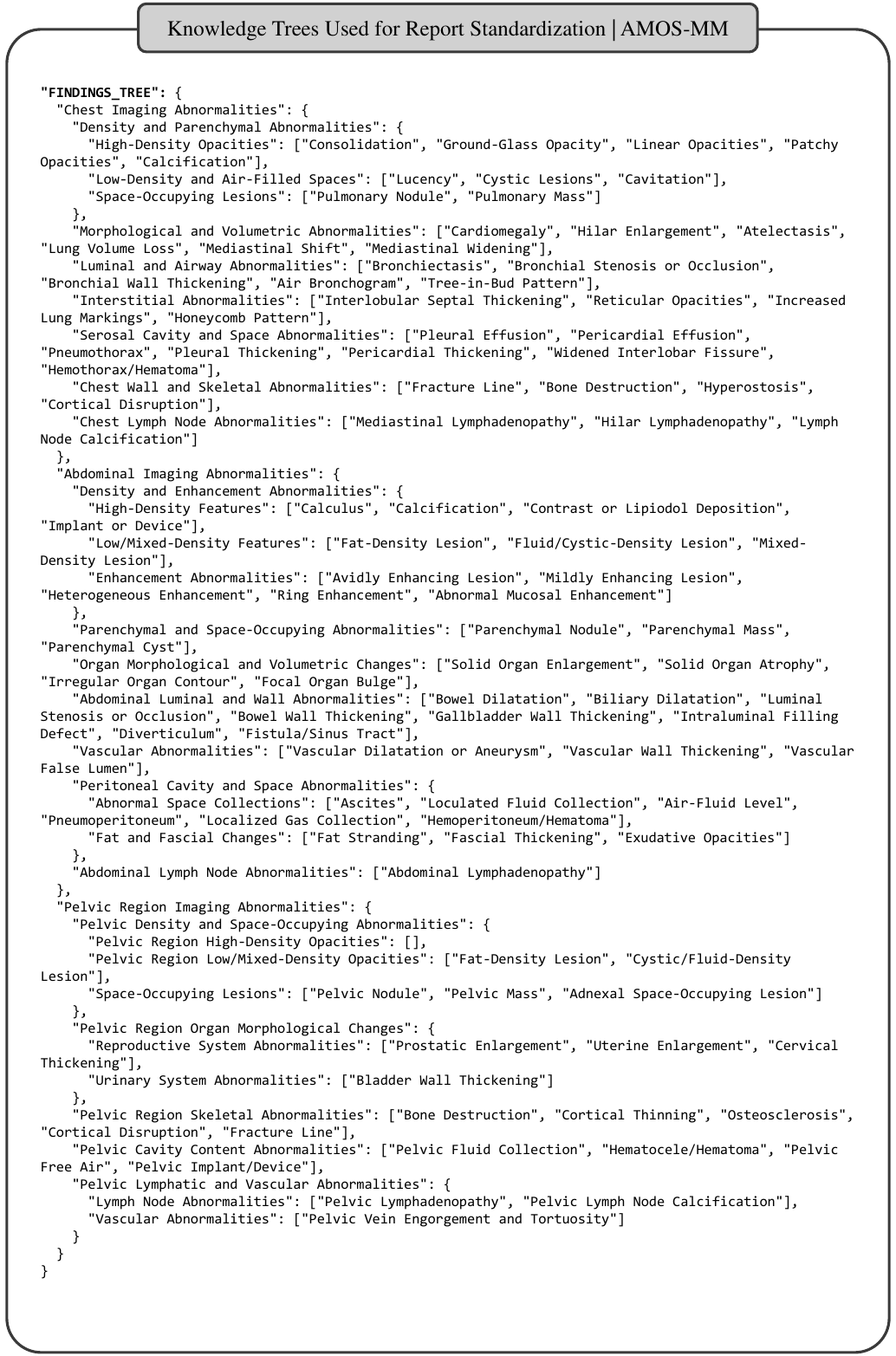}
  \caption{Knowledge trees for AMOS-MM~\citep{amos}.}
  \label{fig:20}
\end{figure}

\begin{figure}
  \centering
  \includegraphics[width=1\linewidth]{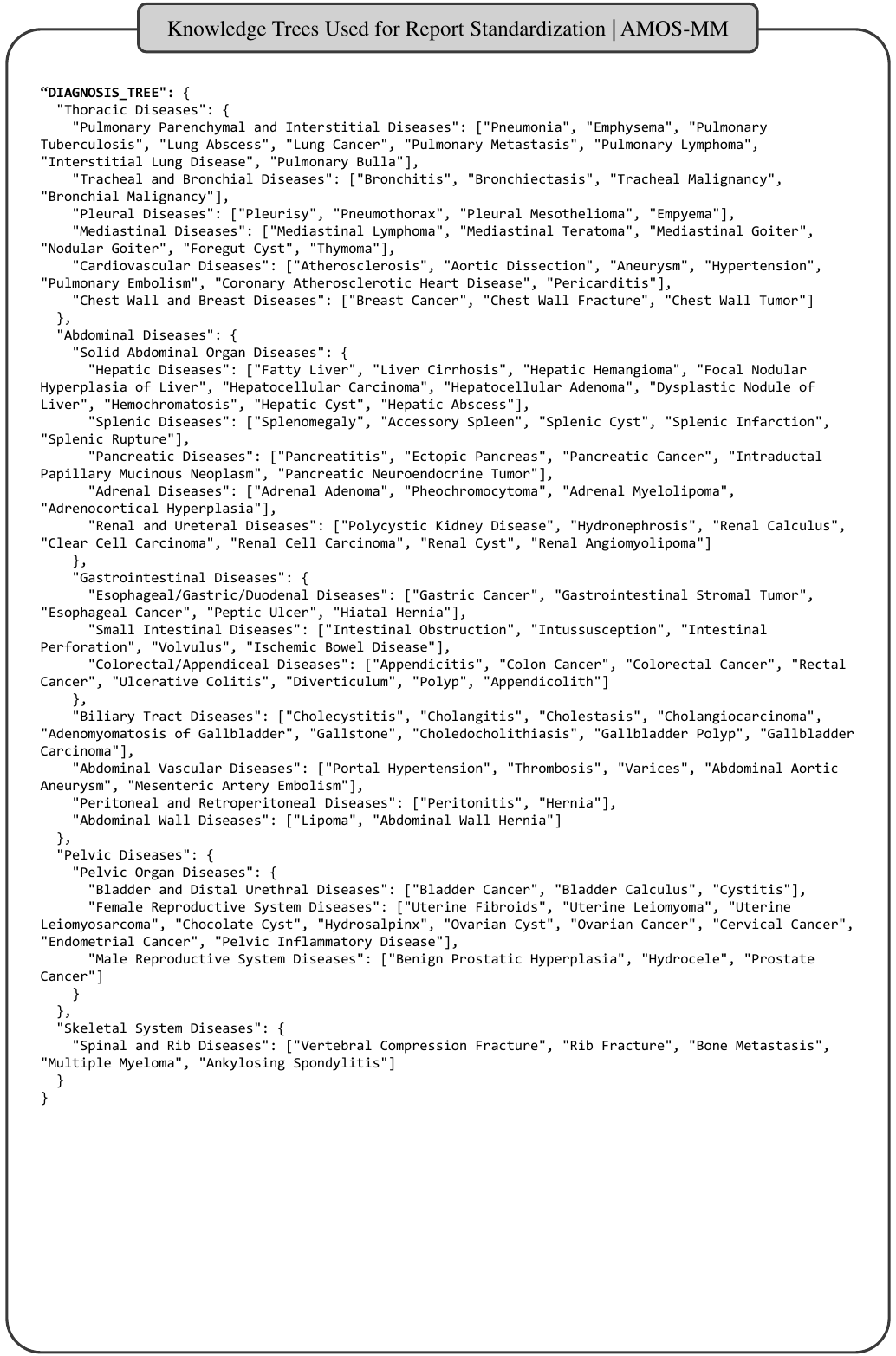}
  \caption{Knowledge trees for AMOS-MM~\citep{amos}.}
  \label{fig:21}
\end{figure}

\begin{figure}
  \centering
  \includegraphics[width=1\linewidth]{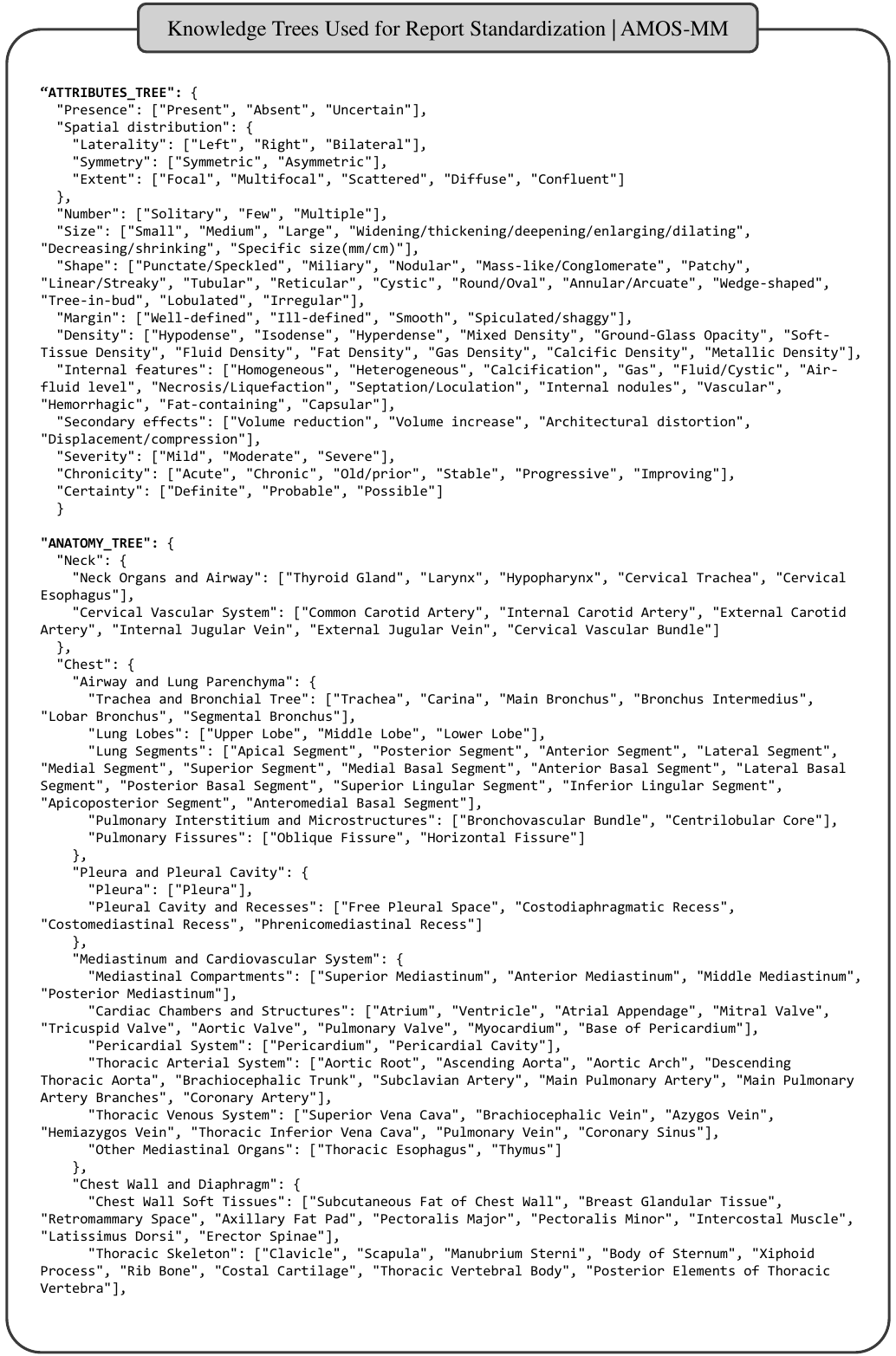}
  \caption{Knowledge trees for AMOS-MM~\citep{amos}.}
  \label{fig:22}
\end{figure}

\begin{figure}
  \centering
  \includegraphics[width=1\linewidth]{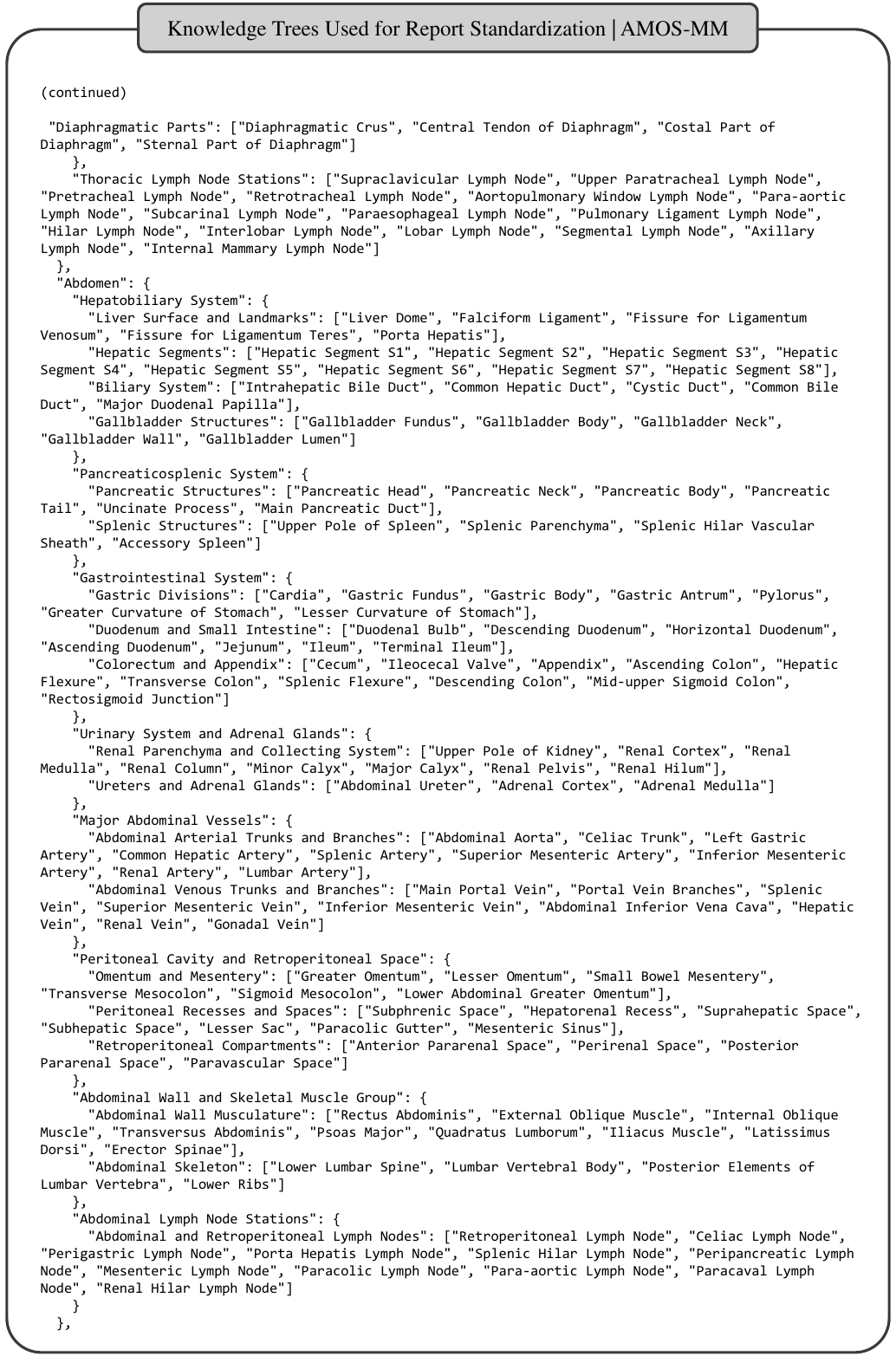}
  \caption{Knowledge trees for AMOS-MM~\citep{amos}.}
  \label{fig:23}
\end{figure}

\begin{figure}
  \centering
  \includegraphics[width=1\linewidth]{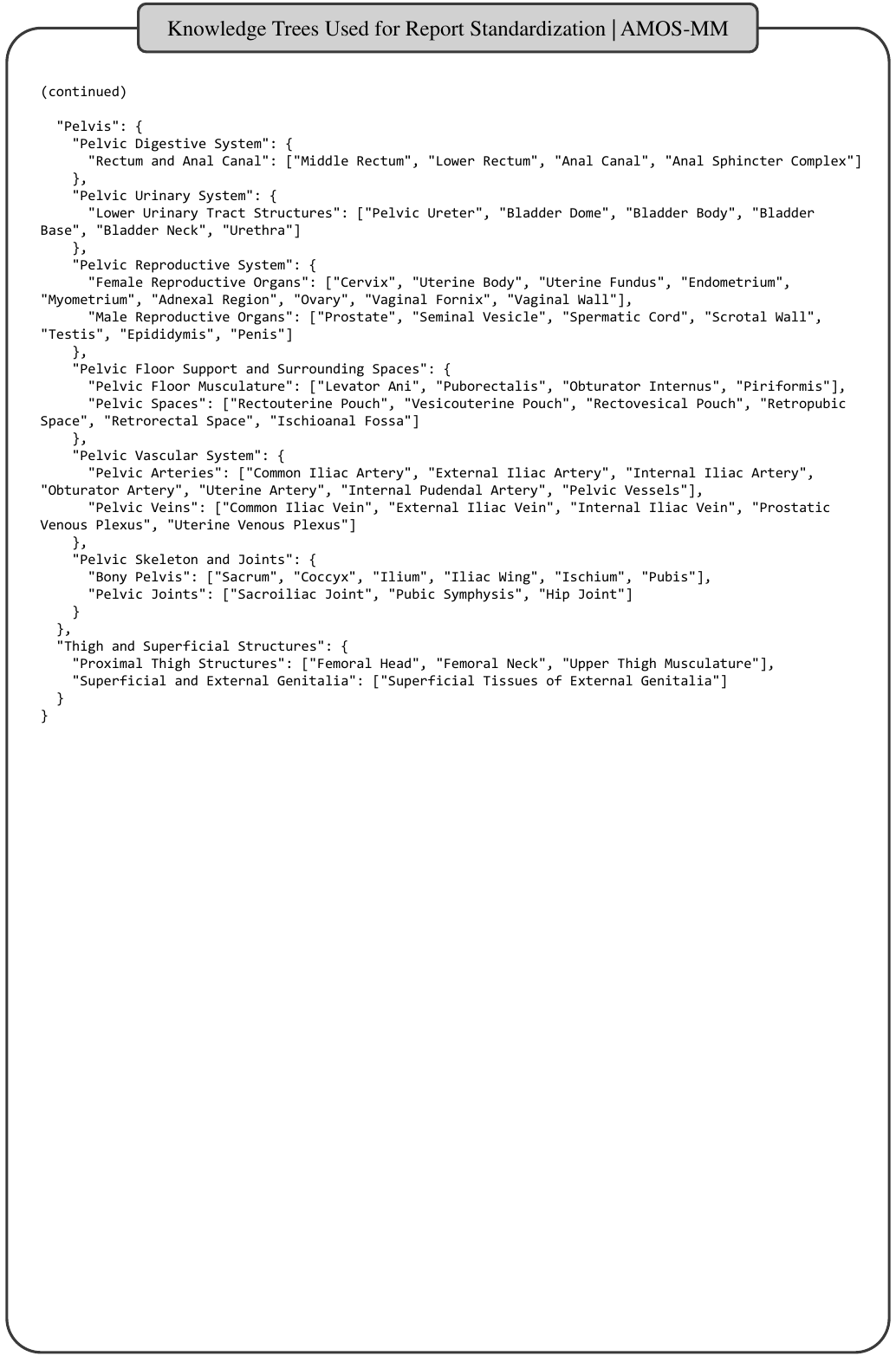}
  \caption{Knowledge trees for AMOS-MM~\citep{amos}.}
  \label{fig:24}
\end{figure}

\begin{figure}
  \centering
  \includegraphics[width=1\linewidth]{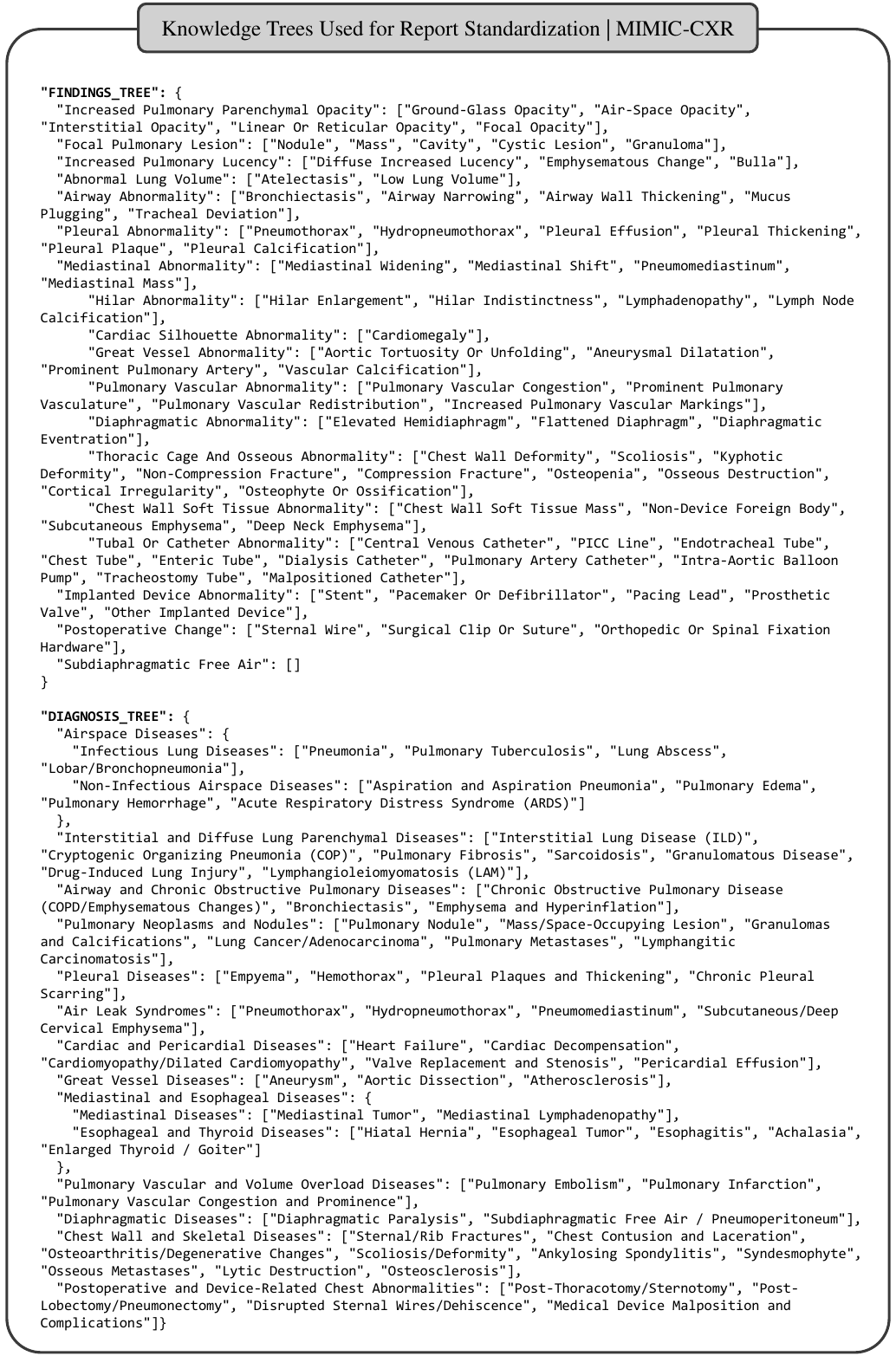}
  \caption{Knowledge trees for MIMIC-CXR~\citep{mimiccxr}.}
  \label{fig:18}
\end{figure}

\begin{figure}
  \centering
  \includegraphics[width=1\linewidth]{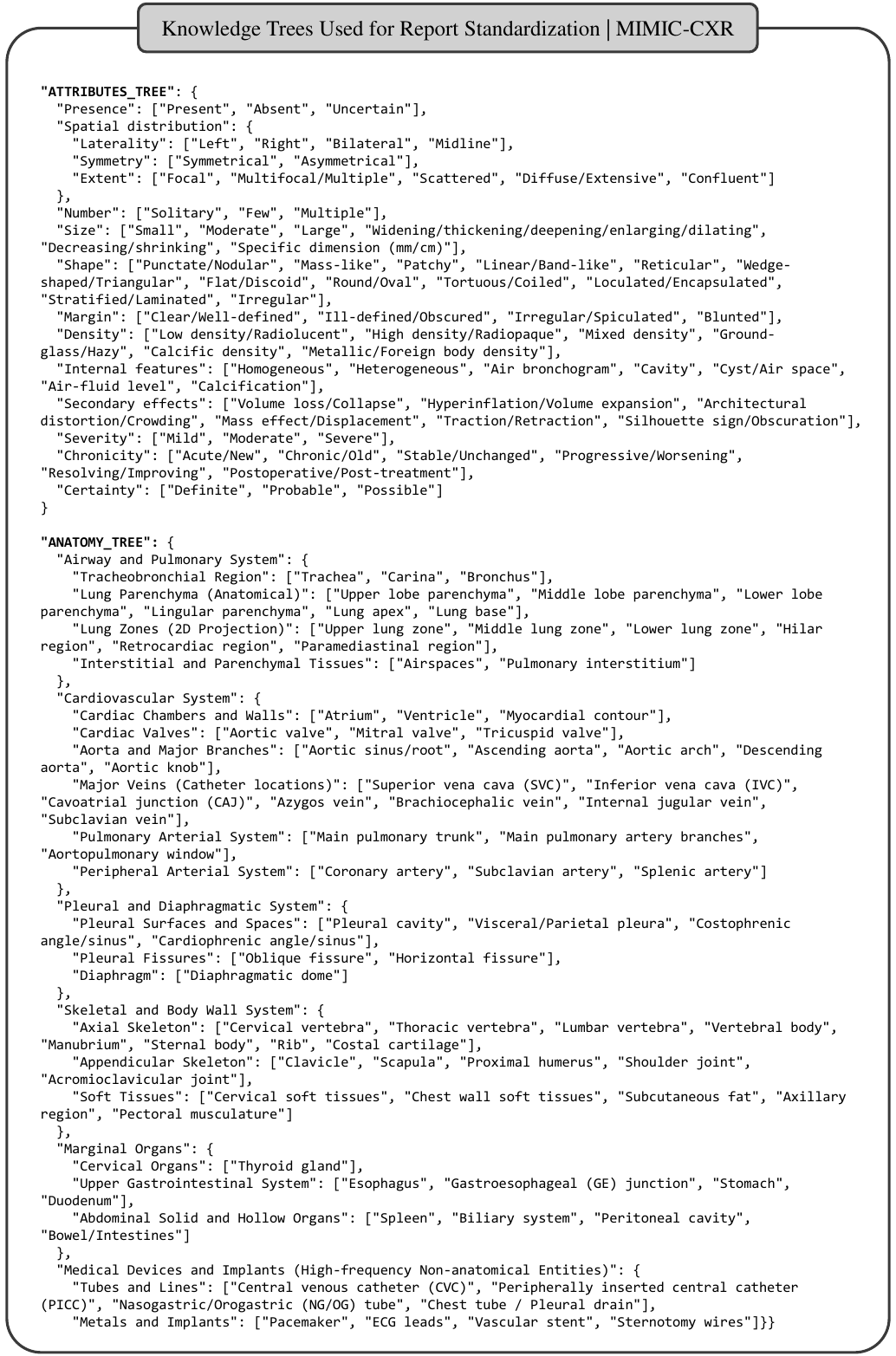}
  \caption{Knowledge trees for MIMIC-CXR~\citep{mimiccxr}.}
  \label{fig:19}
\end{figure}

Figures~\ref{fig:1},~\ref{fig:2},~\ref{fig:3},~\ref{fig:4},~\ref{fig:5},~\ref{fig:6},~\ref{fig:7},~\ref{fig:8},~\ref{fig:9},~\ref{fig:10},~\ref{fig:11},~\ref{fig:12} illustrate prompts used for QA construction. Figures~\ref{fig:13},~\ref{fig:14} show knowledge trees for CTRG-Brain~\citep{ctrg}. Figures~\ref{fig:15},~\ref{fig:16},~\ref{fig:17} show knowledge trees for CT-RATE~\citep{ctrate}. Figures~\ref{fig:20},~\ref{fig:21},~\ref{fig:22},~\ref{fig:23},~\ref{fig:24} show knowledge trees for AMOS-MM~\citep{amos}. Figures~\ref{fig:18},~\ref{fig:19} show knowledge trees for MIMIC-CXR~\citep{mimiccxr}.

\end{document}